\documentclass[11pt]{article}

\usepackage[preprint]{acl}

\usepackage{times}
\usepackage{latexsym}

\usepackage[T1]{fontenc}

\usepackage[utf8]{inputenc}

\usepackage{microtype}

\usepackage{inconsolata}

\usepackage{graphicx}

\usepackage{amsmath}
\usepackage{amssymb}
\usepackage{mathtools}
\usepackage{amsthm}
\usepackage{multirow}
\usepackage{array}
\usepackage{booktabs}
\usepackage[table]{xcolor}
\usepackage{colortbl}
\usepackage{arydshln}
\usepackage{tcolorbox}
\usepackage{tabularx}
\usepackage{subcaption}
\definecolor{myred}{HTML}{F67280}
\definecolor{myblue}{HTML}{31ACD0}
\definecolor{textgreen}{HTML}{508D69}
\definecolor{textred}{HTML}{FF5555}
\definecolor{frameblue}{HTML}{9DBDFF}
\hypersetup{
    colorlinks=true,
    linkcolor=myred,     
    citecolor=myblue,      
}

\usepackage[capitalize,noabbrev]{cleveref}

\theoremstyle{plain}

\theoremstyle{definition}

\theoremstyle{remark}

\title{From Perception to Simulation:  Unveiling and Empowering \\ Mental Spatial Dynamics in MLLMs}

\author{Jiachun Li\textsuperscript{1,2}, Zhuoran Jin\textsuperscript{1,2},  \textbf{Kang Liu\textsuperscript{1,2}, Jun Zhao\textsuperscript{1,2}\footnotemark[1]} \\ \textsuperscript{1}School of Artificial Intelligence, University of Chinese Academy of Sciences \\ \textsuperscript{2}The Key Laboratory of Cognition and Decision Intelligence for Complex Systems, \\ Institute of Automation, Chinese Academy of Sciences \\
\footnotesize{\texttt{\{jiachun.li, zhuoran.jin, kliu, jzhao\}@nlpr.ia.ac.cn }}}

\begin{document}
\maketitle
\renewcommand{\thefootnote}{\fnsymbol{footnote}}
\footnotetext[1]{Corresponding authors.}
\begin{abstract}
    
 Spatial reasoning is a fundamental component of world understanding for Multimodal Large Language Models (MLLMs). While existing research has advanced the perception of spatial relations in static scenes, the domain of dynamic spatial simulation remains relatively unexplored. To bridge this gap, we introduce MentalGrid, a benchmark focused on spatial reasoning tasks that require models to simulate evolving spatial states without intermediate visual feedback. 
    Extensive evaluation on our benchmark reveals a significant deficiency in the spatial simulation capabilities of existing models. Specifically, we find that performance is constrained by perceptual limitations in simple simulation scenarios and hindered by inadequate multi-step reasoning as complexity increases.
    To address these challenges, we propose a three-stage curriculum learning framework utilizing 16k specialized spatial reasoning samples. This approach yields SpatialWalker, which achieves a 33.8\% average improvement over its base model, surpassing proprietary models like Gemini-3-Flash in spatial simulation. Furthermore, SpatialWalker demonstrates robust generalization, maintaining a 4.9\% average gain on out-of-domain benchmarks, demonstrating the pivotal role of spatial simulation as a fundamental building block for broader spatial intelligence.
    
\end{abstract}

\section{Introduction}
Spatial reasoning is the ability of an agent to perceive, understand, and mentally manipulate spatial relationships between objects \citep{spatialcontext,spatialtree,spatial_survey}. It constitutes a fundamental component of how multimodal large language models (MLLMs) understand the physical world. Despite recent advances that enable models to achieve impressive performance on a wide range of visual reasoning tasks \citep{mmmu, visualpuzzle, mmr-v}, their spatial reasoning capabilities remain limited \citep{thinkinspace, spatialladder, spatialtree}. This gap has motivated growing interest in systematic investigations of spatial reasoning in MLLMs.
\begin{figure}[t]
    \centering
    \includegraphics[width=0.95\linewidth]{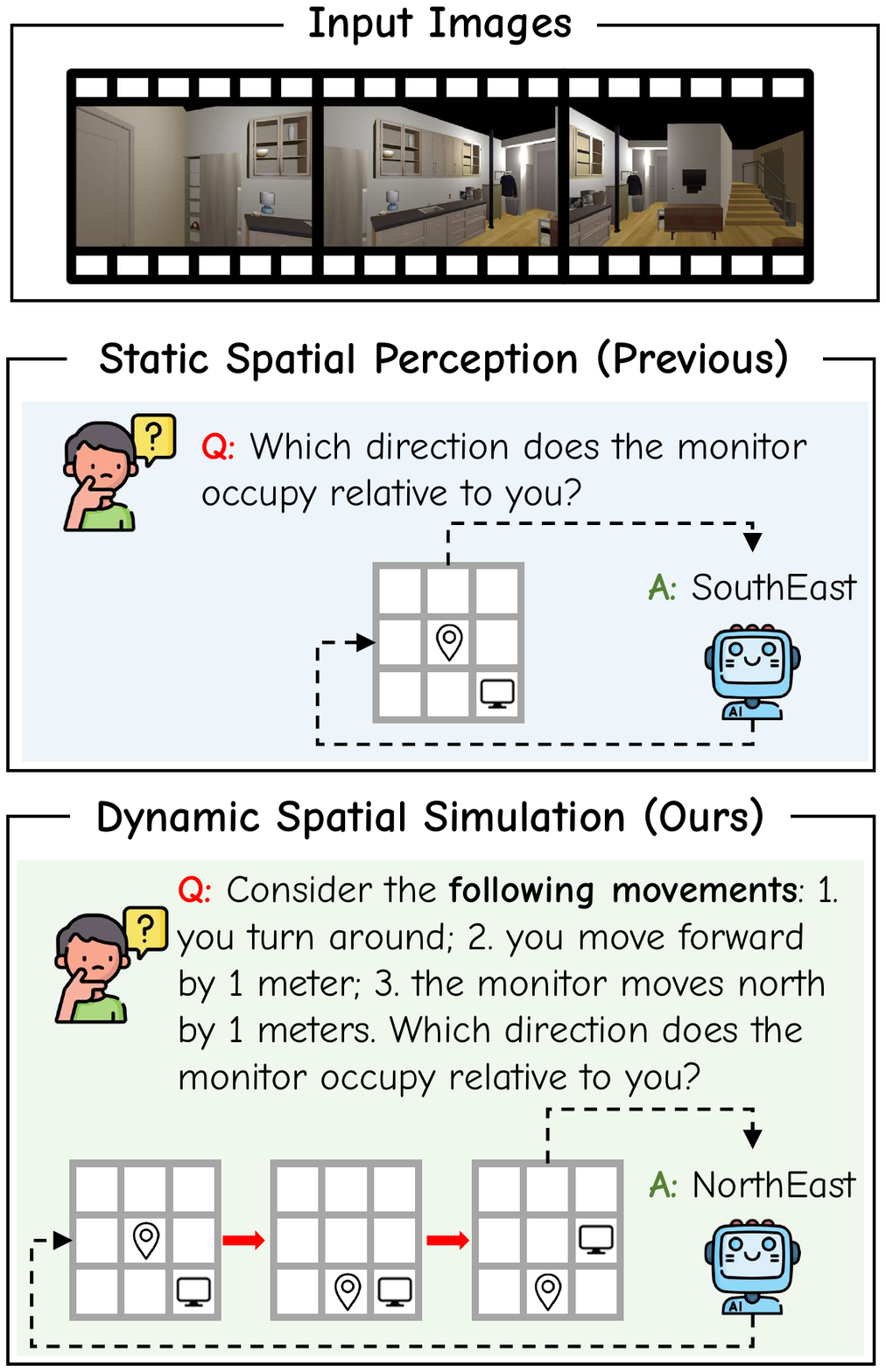}
    \caption{A comparison of two types of spatial reasoning capabilities. Arrows indicate the models’ reasoning processes.} \label{fig:intro}
  \vspace{-15pt}
\end{figure}

Recent work has primarily studied spatial reasoning by focusing on models’ spatial perception capabilities \citep{spatial-mllm, embodied-r, sims-v}.
As illustrated by the first example in Figure \ref{fig:intro}, these works typically assume static environments, in which the spatial states of objects and agents remain fixed \citep{ViewSpatialbench,thinkinspace,mmsibench}. 
Under this setting, models can solve the task by performing a single-pass perception over the input images or videos to localize objects.
However, real-world environments are inherently dynamic, where the spatial positions of objects and agents continuously change over time.
As shown in the second example of Figure \ref{fig:intro}, reasoning in such settings requires models to mentally simulate the evolving spatial state of the environment (e.g., the changing direction between the monitor and agent), a capability we refer to as \textbf{spatial simulation}. While existing studies have explored mental simulation, they predominantly focus on isolated objects \citep{ominispatial, stare} or restricted viewpoints \citep{ViewSpatialbench, mindcube}, leaving this capability largely underexplored.

\begin{figure*}[t]
  \begin{center}
    \centering
    \includegraphics[width=\linewidth]{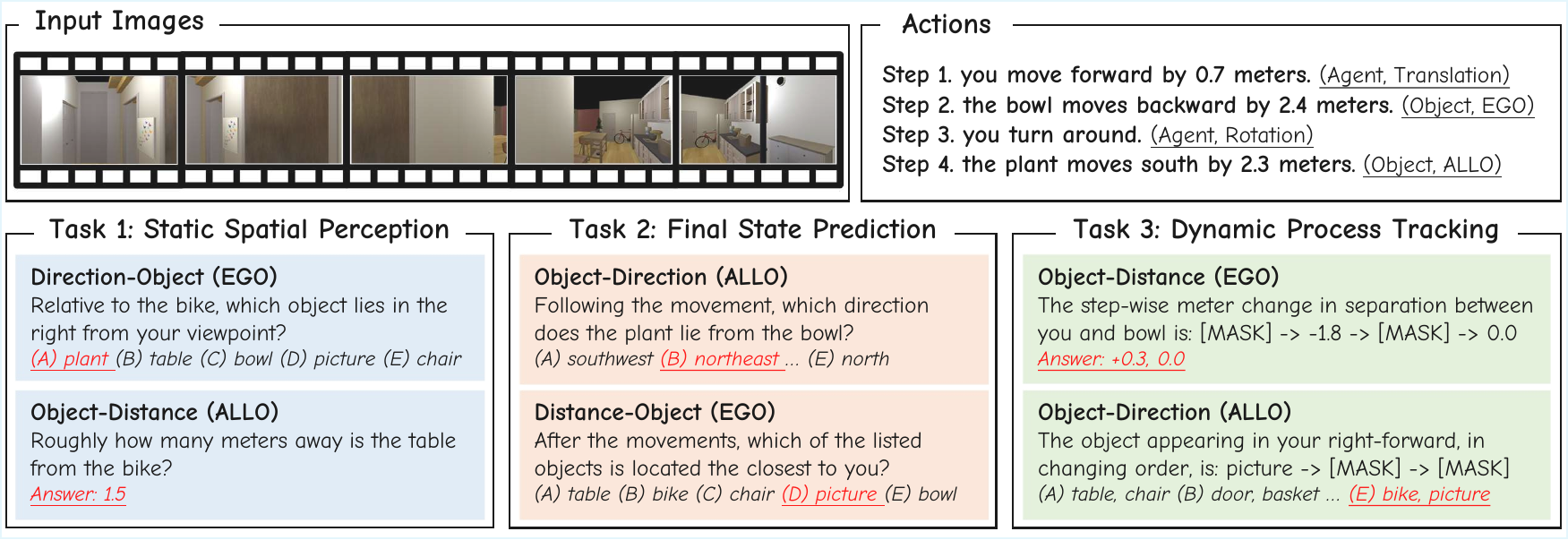}
    \caption{Examples in our MentalGrid benchmark.}
    \label{fig:bench}
  \end{center}
  \vspace{-15pt}
\end{figure*}

In cognitive neuroscience, spatial simulation is supported by internal representations like grid cells \citep{gridcell1,gridcell2,gridcell3}, which enable humans to mentally track their spatial state without continuous visual feedback.
To investigate whether MLLMs exhibit a similar capacity, we introduce \textbf{MentalGrid}, a spatial reasoning benchmark featuring 3,356 question-answer pairs that evaluates models’ ability to predict evolving spatial states without intermediate visual feedback.
As illustrated in Figure \ref{fig:bench}, our benchmark goes beyond static spatial perception by integrating language-guided action sequences into spatial reasoning. It introduces final state prediction and dynamic process tracking tasks, requiring models to predict final spatial states such as distances and directions, as well as to track state changes throughout the movements. 

Our extensive evaluation of 12 state-of-the-art MLLMs on MentalGrid reveals that existing models struggle with spatial simulation, while the best-performing Gemini-3.1-Pro achieves only around \textbf{58}\% accuracy, remaining approximately \textbf{16\%} behind human performance.
Besides, for open-source models, spatial simulation performance is substantially worse than spatial perception.
We further analyze the bottlenecks in the model's spatial simulation performance, finding that perceptual limitations constrain performance for simple simulation steps, while inadequate multi-step reasoning becomes the main limitation as complexity increases.

To mitigate these limitations, we design a three-stage curriculum learning framework based on 16k curated spatial reasoning samples. This framework starts with foundational perceptual grounding, advances to process-aware simulation training to internalize spatial transitions, and culminates in reinforcement learning-based optimization to refine fine-grained simulation trajectories.
This approach yields \textbf{SpatialWalker}, a 3B model that achieves a \textbf{33.8\%} average improvement over its base, surpassing proprietary systems like Gemini-3-Flash. Furthermore, SpatialWalker exhibits robust generalization with \textbf{4.9\%} average performance gains on out-of-domain benchmarks, underscoring the role of spatial simulation as a fundamental building block for broader spatial intelligence.

Our contributions are as follows: 
\textbf{(1) MentalGrid Benchmark:} a novel benchmark for dynamic spatial simulation that requires tracking evolving states across action sequences without intermediate visual feedback. \textbf{(2) Curriculum Learning Framework:} a three-stage strategy, derived from our diagnostic findings, that evolves from perceptual grounding to spatial simulation trajectory optimization. \textbf{(3) SpatialWalker Model:} a 3B model that outperforms its base by \textbf{33.8\%}, surpasses Gemini-3-Flash, and achieves an average \textbf{4.9\%} out-of-domain gains. The relevant data and code are publicly available at: \href{https://github.com/BugMakerzzz/MentalGrid}
{https://github.com/BugMakerzzz/MentalGrid}.

\section{MentalGrid Benchmark}
\subsection{Overview}
We introduce MentalGrid, a multi-image spatial reasoning benchmark designed to evaluate the spatial simulation capabilities of MLLMs. MentalGrid comprises 3,356 questions over 1,100 simulated images. These images are generated from 349 distinct scenes with the Habitat-Sim simulator \citep{habitat}, representing a diverse set of real-world environments commonly encountered in daily life (e.g., offices, homes). 
As shown in Figure \ref{fig:bench}, each question in MentalGrid consists of image-based inputs specifying the initial spatial state, along with language-defined action sequences that update the environment state.
Given these inputs, we require models to perform simulations over the dynamic spatial environment, including inter-object distances and relative directions, without access to intermediate visual observations. We report dataset statistics and comparisons with other benchmarks in the Appendix \ref{append:benchmark}.

\subsection{Task Formulation}
We formally define a spatial simulation trajectory $\mathcal{T}$ as the transition from an initial environment state $\mathcal{S}_0$ to a final state $\mathcal{S}_n$ following a sequence of actions $\mathcal{A}$:
\begingroup
\begin{equation}\label{eq:1}
\footnotesize{
   \begin{aligned}
    \mathcal{T} = \mathcal{S}_0 \xrightarrow{\mathcal{A}} \mathcal{S}_f = \mathcal{S}_0\xrightarrow{a_1} \mathcal{S}_1 \xrightarrow{a_2} \cdots \xrightarrow{a_n} \mathcal{S}_n
    \end{aligned}}
\end{equation}
\endgroup
where $\mathcal{A} = \{a_1 \cdots a_n\}$, $a_i$ denotes the action in step $i$, $n$ is the length of $\mathcal{A}$ and $\mathcal{S}_i$ represents the intermediate environment state after executing action $a_i$. 
Given $\mathcal{A}$ and the input images $\mathcal{I}$, the model is required to correctly predict any intermediate state $\mathcal{S}_i$ in $\mathcal{T}$. Based on the index $i$, we define three task categories:
\textbf{(1) Static Spatial Perception (SSP):} When $i = 0$, the task degenerates into a static spatial perception setting, where the model only needs to interpret $\mathcal{S}_0$ from $\mathcal{I}$ without any action execution. This setting serves as a baseline to compare against the two spatial simulation tasks in identical scenes.
\textbf{(2) Final State Prediction (FSP):} When $i = n$, the model is required to predict the spatial state $\mathcal{S}_n$ after the execution of the whole action sequence $\mathcal{A}$, evaluating its ability to simulate the final outcome of multi-step spatial transformation.
\textbf{(3) Dynamic Process Tracking (DPT):} Beyond final outcome prediction, we also evaluate whether a model can continuously simulate the spatial dynamics throughout the action sequence.
Specifically, models are required to track arbitrary intermediate states $\mathcal{S}_i$ for $0 < i < n$, as well as the transitions $\Delta(\mathcal{S}_i, \mathcal{S}_{i+1})$ between them. 

In concrete questions, we instantiate the state $\mathcal{S}_i$ as a set of object coordinates $\mathcal{P}_i$, the agent position $\mathbf{v}_i$, and the agent orientation vector $\mathbf{o}_i$:
\begingroup
\begin{equation}\label{eq:2}
  \footnotesize{ \begin{aligned}
    \mathcal{S}_i &= (\mathcal{P}_i, \mathbf{v}_i,  \mathbf{o}_i), \quad \mathbf{v}_i, \mathbf{o}_i  \in \mathbb{R}^2  \\
    \mathcal{P}_i &= \{ \mathbf{p}_j \}_{j=1}^M, \quad \mathbf{p}_j \in \mathbb{R}^2 
    \end{aligned}}
\end{equation}
\endgroup
where $M$ denotes the number of objects, and $j$ indexes individual objects. We design different types of queries to evaluate models’ understanding of these coordinates and orientations (see Figure \ref{fig:bench} for examples): 
\textbf{(1) Distance-Object:} Identifying the object that is closest to or farthest from a target object or the agent.
\textbf{(2) Object-Distance:} Measuring distance or distance changes between objects, or between an object and the agent. 
\textbf{(3) Direction-Object:} Identifying the object located in a specified direction relative to a target object or the agent. \textbf{(4) Object-Direction:} Measuring the relative direction or direction changes between objects, or between an object and the agent.
In addition, for each question and action, we define two reference frames: allocentric (ALLO) and egocentric (EGO). The ALLO reference frame adopts a fixed global direction (e.g., North), while the EGO reference frame is defined by the agent’s current orientation.

\begin{table*}[t]
\centering

\resizebox{0.9\linewidth}{!}{
\begin{tabular}{lccccccc}
\toprule
\multirow{2}{*}{\textbf{Model}} & \textbf{SSP} & \multicolumn{2}{c}{\textbf{FSP}} & \multicolumn{2}{c}{\textbf{DPT}} & \multicolumn{2}{c}{\textbf{Overall}} \\ 
\cmidrule(r){2-2} \cmidrule(r){3-4} \cmidrule(r){5-6} \cmidrule(r){7-8} 
&  Acc (\%) \textcolor{textgreen}{$\boldsymbol{\uparrow}$}   & Acc (\%) \textcolor{textgreen}{$\boldsymbol{\uparrow}$}  & Drop (\%) \textcolor{textred}{$\boldsymbol{\downarrow}$}  & Acc (\%) \textcolor{textgreen}{$\boldsymbol{\uparrow}$}   & Drop (\%) \textcolor{textred}{$\boldsymbol{\downarrow}$} & Acc (\%) \textcolor{textgreen}{$\boldsymbol{\uparrow}$}  & Drop (\%) \textcolor{textred}{$\boldsymbol{\downarrow}$} \\
\midrule
Human Level$^\dagger$ & 71.80 & 76.95 & -7.17 & 76.00 & -5.85 & 74.73 & -6.51\\
\midrule
 \multicolumn{8}{c}{\cellcolor{myblue!10!white} \textit{Proprietary Models}} \\
  Gemini-3.1-Pro$^\dagger$ & \textbf{64.00} & \textbf{65.40} & \textbf{-2.19} & 47.40 & 25.94 & \textbf{58.93} & 11.88 \\ 
  GPT-5$^\dagger$          & 53.60 & 38.40 & 28.36 & \textbf{59.70} & \textbf{-11.38} & 50.57 &  8.49 \\
 Gemini-3-Flash & 45.28 & 44.23 & 2.32 & 46.27 & -2.19 & 45.24 & \textbf{0.06} \\
GPT-4o & 38.84 & 32.26 & 16.94 & 20.49 & 47.23 & 30.85 & 32.09 \\
\midrule 
 \multicolumn{8}{c}{\cellcolor{myblue!10!white} \textit{Spatial Models}} \\
 Spatial-SSRL-7B & 28.79 & 21.92 & 23.87 & 17.35 & 39.73 & 22.87 & 31.80 \\
 SpatialLadder-3B & 21.29 & 18.34 & 13.87 & 18.15 & 14.73 & 19.30 & 14.30 \\
  SpatialThinker-3B & 23.21 & 18.43 & 20.62 & 12.09 & 47.92 & 18.10 & 34.27 \\
 SpaceR-7B &  32.01 & 24.15 & 24.57 & 18.41 & 42.51 & 25.07 & 33.54 \\
\midrule
  \multicolumn{8}{c}{\cellcolor{myblue!10!white} \textit{Open-Source Models}} \\
Qwen3-VL-32B     & \underline{40.86} & \underline{41.18} & \underline{-0.78} & \underline{25.61} & 37.31 & \underline{36.19} & 18.26 \\ 
InternVL3.5-38B  & 36.67 & 31.05 & 15.33 & 23.98 & 34.61 & 30.78 & 24.97 \\
Gemma3-12B & 31.02 & 29.62 & 4.52 & 22.43 & 27.68 & 27.85 & 16.10 \\
Qwen2.5-VL-7B & 25.46 & 24.99 & 1.84 & 23.22 & \underline{8.80} & 24.59 & \underline{5.32} \\

\bottomrule
\end{tabular}
}

\caption{Evaluation on MentalGrid. The best result among all models is indicated in \textbf{bold} and the best result among open-source models is \underline{underlined}. $^\dagger$ indicates results on the MentalGrid-tiny set.}
\label{tab:benchmark}
\vskip -0.1in
\end{table*}

\subsection{Data Curation Pipeline} \label{sec:3.3}
We develop a sophisticated data curation pipeline to automatically generate high-quality question-answer pairs.

\textbf{Environment Synthesis. }
Firstly, we synthesize environments using the Habitat-Sim simulator, which supports scalable simulation of diverse 3D scenes with accurate ground-truth spatial annotations and provides precise control over object and agent states. 
We collect 349 different scenes from the ReplicaCAD \citep{replica}, ArchitecTHOR, iTHOR, ProcTHOR, and RoboTHOR \citep{ai2thor}. 
Each scene contains a unique object layout that defines the initial state $\mathcal{S}_0$.
To obtain visual inputs, we position the camera at a designated location and capture multiple images by rotating it within the scene.
The camera orientation in the first image is used to define the fixed global north and the agent’s initial forward orientation $o_0$.

\textbf{Query Construction. }
For each scene, we randomly select either two objects or one object and the agent as query targets, and populate hand-crafted templates to generate four types of queries across the three tasks. To accommodate different query formulations, Object-Distance queries are designed as open-ended questions, while the others adopt a five-choice multiple-choice format. 

\textbf{Action Specialization. }
After generating the queries, we randomly sample an action sequence $\mathcal{A}$ to construct complete FSP and DPT questions. Here, each $a_i \in \mathcal{A}$ is formulated as two types of function mapping $f$:
\begingroup
\begin{equation}\label{eq:3}
 \footnotesize{  \begin{aligned}
     f_t(\mathbf{p}, \theta, d)
&= \mathbf{p} + d \cdot \begin{pmatrix}\cos \theta \\\sin \theta \end{pmatrix} \\
     f_r(\mathbf{o}_i, \theta)
&= \begin{pmatrix} \cos \theta & -\sin \theta \\ \sin \theta & \cos \theta \end{pmatrix} \mathbf{o}_i
    \end{aligned} }
\end{equation}
\endgroup
where $f_t$ denotes the translation transformation and $f_r$ denotes the rotation transformation. The variable $\mathbf{p}$ represents the position, $d$ denotes the translation distance, and $\theta$ denotes the direction.
We randomly sample action steps, transformation functions, distances, and directions to generate $\mathcal{A}$, while ensuring that most actions in $\mathcal{A}$ modify the states of queried targets. 


\textbf{Answer Generation.} 
Finally, given a query and an action sequence, we compute the full state trajectory $\mathcal{T}$ from the initial state $\mathcal{S}_0$, from which the ground-truth answer to the query can be derived. We then generate negative options by randomly sampling incorrect directions or objects, completing the construction of each question-answer pair. 

\textbf{Quality Control.}
While the preceding steps ensure question validity, we implement a multi-stage refinement process to maximize data quality:
\textbf{(1) Filtering Small Objects:} Objects occupying less than 0.1\% of the image pixels are excluded to mitigate perceptual challenges such as occlusion and low resolution.
\textbf{(2) Removing similar objects: } Since our tasks require each object to be unique, we remove scenes containing duplicated or semantically similar object categories.
\textbf{(3) Resolving Spatial Overlap:} To prevent ambiguity caused by discarding height information in 2D coordinate projections, we exclude target objects whose spatial distances fall below a predefined threshold to avoid vertical overlap.
We provide additional details and prompts in Appendix \ref{append:data_curation}.

\section{Spatial Simulation Evaluation} \label{sec:4}


\subsection{Evaluation Setup} \label{sec:4.1}

\textbf{Evaluation Models.} We conduct a comprehensive evaluation of 12 diverse MLLMs. For proprietary models, we consider Gemini-3.1-Pro \citep{gemini3.1}, GPT-5 \citep{gpt-5}, Gemini-3-Flash \cite{gemini3} and GPT-4o \citep{gpt-4o}. For open-source models, we evaluate Qwen3-VL-32B \citep{qwen3-vl}, InternVL3.5-38B \citep{internvl-3.5}, Gemma3-12B \citep{gemma3} and Qwen2.5-VL-7B \citep{qwen2.5-vl}. For spatial models, we select SpatialLadder-3B \citep{spatialladder},  SpatialThinker-3B \citep{spatialthinker}, Spatial-SSRL-7B \citep{spatialssrl} and SpaceR-7B \citep{spacer}. 

\textbf{Human Performance.} We extract 50 examples from each task to form a tiny test set. From it, we sample 30 questions at a time and assign them to one of five students, then calculate the average human performance.



\textbf{Metrics.} We use Accuracy (Acc) for multiple-choice questions and Mean Relative Accuracy (MRA) for open-ended questions \citep{thinkinspace}. MRA averages performance across multiple confidence thresholds $\tau$ based on relative error:
\begingroup
\begin{equation}\label{eq:4}
   \footnotesize{\begin{aligned}
    \mathrm{MRA} = \frac{1}{10} \sum_{\tau \in \mathcal{C}} \mathbb{I}\left( \frac{|\hat{y} - y|}{y} < 1 - \tau \right)
    \end{aligned}}
\end{equation}
\endgroup
where $\hat{y}$ denotes the model's prediction, $y$ denotes the ground truth, and $\mathcal{C} = \{0.5, 0.55, ..., 0.95\}$. 
For brevity, we use ``Accuracy'' to refer to both metrics.
To disentangle errors between spatial simulation and perception, we define a relative drop rate for each task $t \in \{\text{FSP}, \text{DPT}\}$ as $\text{Drop}_t = (\text{Acc}_{\text{SSP}} - \text{Acc}_t) / \text{Acc}_{\text{SSP}}$. A higher drop indicates greater difficulty in internal spatial simulation beyond static perception. Additional details and comprehensive results across a wider range of frontier models on both MentalGrid and MentalGrid-tiny are provided in Appendix~\ref{append:main_eval}.

\subsection{Main Results on MentalGrid}  \label{sec:4.2}
Table \ref{tab:benchmark} shows overall model performance on MentalGrid. Our key observations are as follows:
\textbf{(1) Current models leave significant room for improvement in spatial simulation.} Although advanced models like Gemini-3.1-Pro demonstrate relatively strong spatial simulation performance (58.93 \% accuracy), Table \ref{tab:benchmark} shows that they still lag behind human performance by \textbf{16\%}. 
Besides, the majority of open-source models perform near the chance level, indicating significant limitations in their spatial simulation capabilities.
\textbf{(2) Open-source models show weaker spatial simulation than spatial perception.}
As shown in Table \ref{tab:benchmark}, most open-source models achieve their best performance on the SSP task, with large positive Drop values. In contrast, proprietary models such as Gemini-3-Flash exhibit relatively consistent performance across different tasks. This suggests that spatial reasoning in open-source models relies heavily on static perception and struggles to generalize when reasoning requires multi-step spatial state simulation.
\textbf{(3) Process modeling is more challenging than outcome prediction.}
In Table \ref{tab:benchmark}, most models exhibit larger performance degradation on DPT than on FSP.
This observation indicates that accurate final-state prediction does not necessarily reflect a faithful understanding of the underlying intermediate spatial dynamics. Consequently, these results highlight the importance of strengthening the entire spatial simulation process, rather than optimizing predictions at isolated states alone.
This finding serves as the foundation for our training framework in \S \ref{sec:5.1}.


\subsection{Impact of Perception Complexity}  \label{sec:4.3}

Spatial simulation first requires perceiving and localizing target objects in the initial scene. We therefore begin by analyzing how perception complexity influences model performance.
Specifically, we group questions by object count and object density, and report accuracy within each group. Results for Qwen2.5-VL-3B are shown in Figure~\ref{fig:perception}, with additional results provided in Appendix~\ref{append:perception_exp}.

\begin{figure}[t]
  \centering
  \begin{subfigure}[t]{0.494\linewidth}
    \centering
    \includegraphics[width=\linewidth]{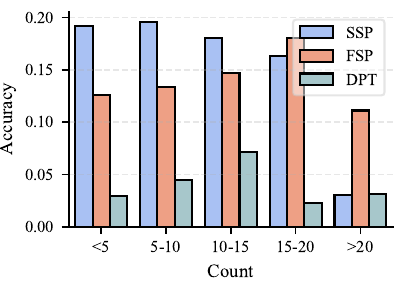}
    \caption{Object Count} \label{fig:obj_cnt}
  \end{subfigure}\hfill
  \begin{subfigure}[t]{0.494\linewidth}
    \centering
    \includegraphics[width=\linewidth]{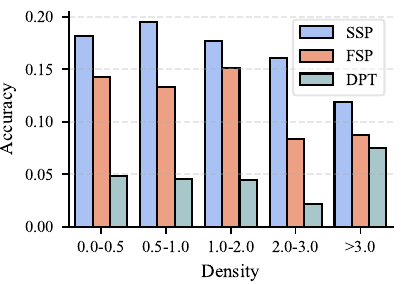}
    \caption{Spatial Density} \label{fig:spa_density}
  \end{subfigure}
  \caption{Evaluation under different perception complexity on Qwen2.5-VL-3B.}
  \label{fig:perception}
  \vspace{-10pt}
\end{figure}

\begin{figure*}[t]
  \centering
  \begin{subfigure}[t]{0.333\linewidth}
    \centering
    \includegraphics[width=\linewidth]{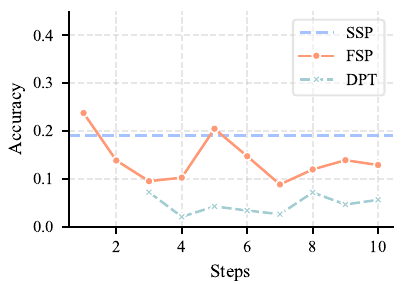}
    \caption{Raw Image} \label{fig:raw_action}
  \end{subfigure}\hfill
  \begin{subfigure}[t]{0.333\linewidth}
    \centering
    \includegraphics[width=\linewidth]{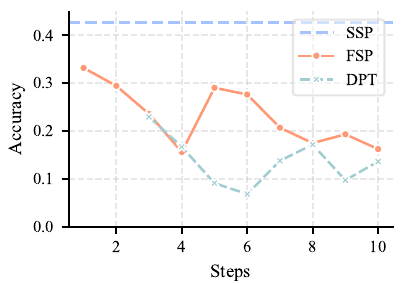}
    \caption{2D Map Image} \label{fig:image_map_action}
  \end{subfigure}\hfill
  \begin{subfigure}[t]{0.333\linewidth}
    \centering
    \includegraphics[width=\linewidth]{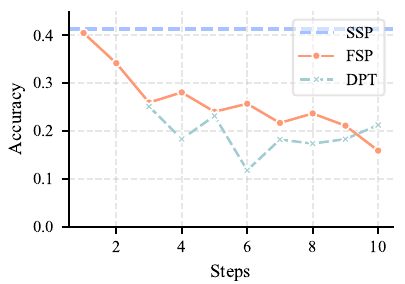}
    \caption{2D Map Text} \label{fig:text_map_action}
  \end{subfigure}
  \caption{Evaluation under different action steps on Qwen2.5-VL-3B.}
  \label{fig:sim_complex}
\vspace{-10pt}
\end{figure*}


For the object count,  Figure \ref{fig:obj_cnt} shows that SSP performance exhibits a sharp decay once the object count in the scene exceeds a critical threshold (e.g., $>20$), uncovering a clear perceptual bottleneck.
However, this degradation is less pronounced for FSP or DPT, suggesting that while increased object scale hampers initial perceptual grounding, spatial simulation performance is relatively less affected by it.
To further quantify environmental complexity, we define Spatial Density ($\rho$) via the Mean Nearest Neighbor Distance \citep{density}:

\vspace{-10pt}
\begingroup
\footnotesize{\begin{equation}\label{eq:5}
  \rho = \frac{1}{M} \sum_{i=1}^{M} \min_{j \neq i} \sqrt{(x_i - x_j)^2 + (y_i - y_j)^2}
\end{equation}}
\endgroup
where $(x_i, y_i)$ is the position of each object. While traditional counts represent global magnitude, a higher $\rho$ captures the local clumping of the environment. 
As shown in Figure \ref{fig:spa_density}, increased density significantly degrades SSP accuracy due to heightened perceptual interference, whereas spatial simulation tasks remain relatively robust to such local crowding.
Overall, spatial simulation is markedly less sensitive to scene complexity than spatial perception. 
This decoupling suggests that: while perceptual grounding constitutes the primary bottleneck for static perception tasks, \textbf{spatial simulation is constrained by additional critical factors beyond perception alone.}

\subsection{Impact of Simulation Complexity}  \label{sec:4.4}
We investigate how the complexity of the simulation process affects model performance.
Given the known limitations of current models in 3D perception \citep{spatialladder}, we introduce two simplified input formats to reduce perceptual confounds in our evaluation: 
\textbf{(1) 2D Map Image:}
We aggregate the coordinates of different objects across multiple scene images to construct a top-down 2D grid map. This input allows models to perform our spatial reasoning tasks without relying on 3D perceptual capabilities such as depth estimation or object detection. 
\textbf{(2) 2D Map Text:}
We further include a JSON-based format that specifies the precise coordinates of all objects in the scene. This input format does not require any visual perception from the model and conveys spatial information purely through language. We demonstrate examples for each format in Figure \ref{fig:input_format}.

We consider the number of action steps and the entropy of action distributions as measures of simulation complexity, and compare model performance under different input formats in Figures~\ref{fig:sim_complex} and~\ref{fig:sim_complex_ent}. From these results (refer to Appendix~\ref{append:simulation_exp} for extended experiments), we establish the following conclusions:
\textbf{(Finding 1) Perceptual Dominance in Low-Complexity Simulations.}
As shown in Figure \ref{fig:image_map_action} and \ref{fig:text_map_action}, providing structured spatial representations (such as 2D map images or text) substantially improves SSP accuracy and yields modest gains for FSP and DSP in small-step regimes.
This indicates that under limited simulation complexity, perceptual grounding still constrains spatial simulation performance.
\textbf{(Finding 2) Long-range Reasoning as the Primary Bottleneck.} However, as the number of action steps increases, FSP performance in Figures~\ref{fig:image_map_action} and \ref{fig:text_map_action} exhibits a pronounced decline, eventually converging to the baseline established by raw inputs in Figure~\ref{fig:raw_action}. This suggests that while explicit spatial grounding can mitigate initial perceptual uncertainty, the compounding difficulty of multi-step reasoning remains the fundamental limiting factor in spatial simulation.
\textbf{(Finding 3) Models Exhibit Minimal Capability for Process-Oriented Spatial Simulation.} 
Even under short action sequences and with explicit coordinate inputs, DPT accuracy remains consistently low, indicating that models struggle to reason about intermediate spatial state transitions.
Such results highlight an intrinsic limitation: current open-source models lack process-oriented spatial simulation capabilities and instead rely on shallow or outcome-focused spatial reasoning strategies.



\begin{table*}[t]
\centering

\resizebox{0.9\linewidth}{!}{
\begin{tabular}{lccccccccc}
\toprule
\multirow{2}{*}{\textbf{Model}} & \multicolumn{4}{c}{\textbf{In-domain}} & \multicolumn{4}{c}{\textbf{Out-of-domain}} & \multirow{2}{*}{\textbf{Overall}} \\ 
\cmidrule(r){2-5} \cmidrule(r){6-9} 
&  SSP  & FSP & DPT & Avg. &  SSP  & FSP & DPT & Avg.   \\
\midrule 
\multicolumn{10}{c}{\cellcolor{myblue!10} \textit{Proprietary Models}} \\

Gemini-3-Flash & 43.48 & 42.44 & 45.37 & 43.73 & \textbf{47.74} & \textbf{46.72} & 47.58 & \textbf{47.34} & 45.24 \\
GPT-4o & 36.81	& 32.23	& 18.68	& 29.50	& 41.61 & 32.30 & 23.11	& 32.74	& 30.85 \\

\midrule
  
\multicolumn{10}{c}{\cellcolor{myblue!10} \textit{Qwen2.5-VL-7B Based Spatial Models}} \\
Qwen2.5-VL-7B & 23.81 & 24.41 & 21.65 & 23.33 & 27.70 & 25.78 & 25.48 & 26.36 & 24.59 \\
Spatial-SSRL-7B & 26.79 & 24.41 & 16.61 & 22.75 & 31.50 & 18.46 & 18.41 & 23.02 & 22.87 \\
SpaceR-7B & 30.09 & 24.10 & 17.37 & 24.02 & 34.63 & 24.22 & 19.90 & 26.54 & 25.07 \\

\midrule
 \multicolumn{10}{c}{\cellcolor{myblue!10} \textit{Qwen2.5-VL-3B Based Spatial Models}} \\
Qwen2.5-VL-3B & 19.43 & 12.12 & 4.35 & 12.16 & 17.11 & 16.22 & 4.92 & 13.06 & 12.54 \\
SpatialThinker-3B & 22.24 & 16.55 & 12.8 & 17.31 & 24.53 & 21.02 & 11.07 & 19.19 & 18.10 \\
SpatialLadder-3B & 21.18 & 17.38 & 18.45 & 19.02 & 21.44 & 19.67 & 17.73 & 19.69 & 19.30 \\
\textbf{SpatialWalker-3B (Ours)} & \textbf{46.51} &	\textbf{51.08}	& \textbf{45.65}	& \textbf{47.79}	& 39.41	& 45.24	& \textbf{49.01}	& 44.35	& \textbf{46.35}  \\
\textit{\textcolor{textgreen}{\quad $\Delta$ \textbf{Improvement}}} & \textcolor{textgreen}{+27.08}	& \textcolor{textgreen}{+38.96}	& \textcolor{textgreen}{+41.30}	& \textcolor{textgreen}{+35.63}	& \textcolor{textgreen}{+22.30}	& \textcolor{textgreen}{+29.02}	& \textcolor{textgreen}{+44.09}	& \textcolor{textgreen}{+31.29}	& \textcolor{textgreen}{+33.81} \\

\bottomrule
\end{tabular}
}
\caption{Evaluation results on our MentalGrid benchmark, the highest performance is indicated in \textbf{bold}.}
\label{tab:main_comp}

\end{table*}

\begin{table*}[t]
\centering

\resizebox{0.9\linewidth}{!}{
\begin{tabular}{lccccccccccc}
\toprule
\multirow{2}{*}{\textbf{Model}}  & \multicolumn{4}{c}{\textbf{MindCube}} & \multicolumn{5}{c}{\textbf{OmniSpatial}} & \multirow{2}{*}{\textbf{STARE}} &  \multirow{2}{*}{\textbf{Overall}}\\ 
\cmidrule(r){2-5} \cmidrule(r){6-10} 

& Among & Around & Rotation & Avg. & DR. & SI. & CL. & PT. & Avg. \\
\midrule

SpatialLadder-3B & \underline{46.33} & \underline{58.00} & 27.00 & \textbf{45.43} & 39.76 & 43.33 & 23.41 & 39.22 & 37.57 & 31.75 & 38.25 \\
SpatialThinker-3B & 40.17 & \textbf{62.00} & 33.50 & 44.10 & 42.14 & 47.33 & 25.00 & \underline{41.35} & 40.05 & 32.33 & \underline{38.83} \\
Spatial-SSRL-7B & 29.83 & 35.20 & \textbf{35.50} & 32.19 & \underline{44.52} & \underline{48.33} & \textbf{26.19} & \textbf{44.03} & \underline{41.06} & \textbf{42.07} & 38.77 \\
Qwen2.5-VL-7B & 27.83 & 38.80 & 33.50 & 31.52 & 30.48 & \textbf{49.00} & 22.62 & 39.75 & 36.20 & \underline{38.53} & 35.42 \\
\midrule

\rowcolor{myblue!5} Qwen2.5-VL-3B & 36.83 & 48.40 & 33.00 & 38.86 & 37.14 & 42.00 & 25.00 & 39.39 & 36.92 & 33.71 & 36.50 \\
\rowcolor{myblue!5} \textbf{SpatialWalker-3B} & \textbf{46.50} & 51.60 & \underline{34.00} & \underline{45.33} & \textbf{55.95} & 43.67 & \underline{25.79} & 40.11 & \textbf{42.79} & 36.14 & \textbf{41.42} \\
\rowcolor{myblue!5} \textbf{\textit{\textcolor{textgreen}{\quad $\Delta$ Improvement}}} & \textcolor{textgreen}{+9.67}	& \textcolor{textgreen}{+3.20} & \textcolor{textgreen}{+1.00} & \textcolor{textgreen}{+6.47}	& \textcolor{textgreen}{+18.81} & \textcolor{textgreen}{+1.67}	& \textcolor{textgreen}{+0.79}	& \textcolor{textgreen}{+0.72}	& \textcolor{textgreen}{+5.87}	& \textcolor{textgreen}{+2.43} & \textcolor{textgreen}{+4.92}
 \\

\bottomrule
\end{tabular}
}

\caption{Evaluation results on other benchmarks. Best results are \textbf{bolded} and second best are \underline{underlined}.}
\label{tab:ood_test}

\vspace{-12pt}
\end{table*}

\section{Spatial Simulation Training}

\subsection{Our Training Framework} \label{sec:5.1}
Guided by the analysis in \S \ref{sec:4}, in this section, we propose a targeted training strategy to address current bottlenecks and enhance spatial simulation capabilities.

\textbf{Training Data Construction.}
We first construct the training data following the pipeline described in \S~\ref{sec:3.3}. 
In particular, when generating training environments, we exclude the ArchitecTHOR and iTHOR datasets, which are reserved for out-of-domain evaluation. 
From the remaining datasets, we generate a total of 16k QA pairs, including 5,368 for the static spatial perception, 5,289 for the final state prediction, and 5,294 for the dynamic process tracking. 
For each question, we prompt GPT-5-mini to think step-by-step and deduce the final answer. Chains of thought (CoT) with accuracy above 0.9 are retained as training trajectories. 

\textbf{Curriculum Learning Framework.} 
Building upon the above training data, we design a three-stage curriculum training framework that systematically improves the model's spatial simulation abilities. 
\textbf{(Stage 1) Perceptual Grounding on Static Spatial Perception:}
Motivated by Finding 1, in the first stage, we perform supervised fine-tuning on the SSP task to establish reliable perceptual grounding.
This stage focuses exclusively on improving object localization and spatial grounding, enabling the model to accurately understand the corresponding spatial relations within static scenes.
\textbf{(Stage 2) Synergistic Learning of Outcome and Process for Spatial Simulation:}
Motivated by Findings 2 and 3, Stage 2 fine-tunes the model on joint FSP and DPT trajectories to address the bottleneck in multi-step state simulation. While FSP incentivizes accurate terminal outcome prediction, DPT enforces rigorous reasoning over intermediate transitions. This synergy enables the model to transition from shallow, outcome-biased heuristics to robust, process-aware spatial simulation.
\textbf{(Stage 3) Fine-Grained State Evolution via Reinforcement Learning:}
To further address the process-oriented deficiencies identified in Finding 3, we introduce a third stage centered on Reinforcement Learning (RL) using the DPT dataset. While Stage 2 establishes a supervised foundation, RL further refines the model's performance on state-evolution trajectories.
We employ GRPO \citep{grpo} for stable policy optimization, utilizing a format reward that enforces a structured reasoning template and an accuracy reward validates the predicted spatial state against the ground truth. By optimizing for this joint objective, the model learns to bridge the gap between static perception and dynamic simulation, fostering a more robust internal representation of spatial state trajectories. 
We report more training details in Appendix \ref{append:training}.

\subsection{Evaluation on MentalGrid}
Table~\ref{tab:main_comp} compares SpatialWalker with state-of-the-art baselines across in-domain and out-of-domain scenes on the MentalGrid benchmark. Our model achieves an overall accuracy of \textbf{46.35\%}, delivering a remarkable \textbf{+33.81\%} improvement over the Qwen2.5-VL-3B backbone. Notably, SpatialWalker not only outperforms existing open-source spatial models but also surpasses proprietary models like Gemini-3-Flash, demonstrating that our multi-stage training effectively resolves the bottleneck in spatial simulation. Furthermore, SpatialWalker exhibits robust generalization, maintaining a \textbf{44.35\%} average accuracy on out-of-domain scenes and achieving the highest accuracy in the challenging DPT task (\textbf{49.01\%}), underscoring its superior ability to generalize spatial transition dynamics to unseen environments. We provide additional comparisons with test-time world-model-based approaches in Appendix~\ref{append:method_eval}, where our model consistently achieves stronger performance.

\subsection{Generalization to Other Benchmarks}

To demonstrate that our model’s performance extends beyond our benchmark, we evaluate its performance across a diverse range of tasks: MindCube \citep{mindcube} tests spatial mental modeling in real-world scenarios, OmniSpatial \citep{ominispatial} evaluates comprehensive spatial reasoning abilities, and STARE \citep{stare} assesses multi-step visual simulation performance (see Appendix \ref{append:ood} for more benchmarks). As shown in Table \ref{tab:ood_test}, our SpatialWalker-3B achieves a superior overall score of \textbf{41.42\%}, consistently outperforming other specialized spatial models. Most notably, our model excels in the dynamic spatial reasoning (DR.) task of OmniSpatial, reaching an \textbf{18.81\%} gain over the base model. These results validate that our approach significantly enhances the model's transferable dynamic spatial reasoning capabilities across diverse scenarios, demonstrating strong generalization to spatial reasoning in real-world environments.
\begin{figure}[t]
    \centering
    \includegraphics[width=0.9\linewidth]{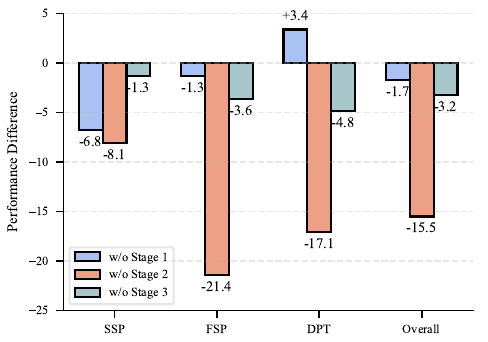}
    \caption{Ablation study results.} \label{fig:ablation}
\vspace{-15pt}
\end{figure}

\subsection{Ablation Study}

Figure \ref{fig:ablation} reveals the critical contribution of each module within our framework.
Stage 2 proves most essential, as its removal causes a 15.5\% overall accuracy drop, validating simulation trajectory training as the cornerstone of dynamic spatial reasoning.
Stage 3 contributes meaningfully with a 3.2\% overall loss, confirming that policy refinement on sequential state transitions is vital for robust simulation.
While excluding Stage 1 causes a 6.8\% degradation in static perception, it leads to a 3.4\% gain in DPT, suggesting a potential trade-off between static perceptual priors and the fluid reasoning required to track evolving process changes. Appendix \ref{append:ablation} provides additional ablation results.

\section{Related Work}
\subsection{Spatial Reasoning in MLLMs}
Spatial reasoning in MLLMs centers on the capacity to perceive, understand, and mentally manipulate spatial relationships between objects. For evaluation, most studies represent object layouts using images or videos to assess models’ understanding of geometric properties and spatial arrangements \citep{thinkinspace, ViewSpatialbench, mmsibench, ominispatial}. For methods, existing works seek to bolster spatial reasoning by enhancing 3D visual perception, primarily focusing on improving perceptual capabilities through the integration of extra visual modules \citep{embodied-r, spatial-mllm}, tool augmentation \citep{vilasr, apc}, or advanced training techniques \citep{spatialladder, spatialthinker}. Our work shifts the focus from static perception toward dynamic spatial simulation.

\subsection{Mental Simulation Evaluation}
Existing research has extensively evaluated the mental simulation capabilities of models in spatial reasoning. Prior studies have investigated geometric manipulations, such as mental rotation and folding \citep{stare, spatialviz}, while others focus on scene perception across varying viewpoints \citep{ViewSpatialbench, mindcube}. Furthermore, several works evaluate model performance in sequential planning and spatial navigation when provided with explicit start and end points \citep{thinkinspace, spatialeval}. However, these studies primarily operate within static environments, leaving a gap in understanding long-horizon transitions and evolving spatial states.

\section{Conclusion}
In this paper, we introduce MentalGrid, a benchmark evaluating the prediction of evolving spatial states without intermediate visual feedback. Our analysis reveals that existing models primarily struggle with spatial simulation. To address this, we develop a three-stage curriculum framework evolving from perceptual grounding to RL-based trajectory optimization. This approach yields SpatialWalker, a 3B model that achieves a \textbf{33.8\%} improvement over its base, surpasses Gemini-3-Flash, and demonstrates robust generalization across out-of-domain benchmarks. Our findings underscore spatial simulation as a fundamental building block for broader spatial intelligence.

\section*{Limitations}
Although our work conducts an in-depth evaluation and proposes training framework for improving MLLMs' spatial simulation abilities, it has several limitations.
First, real-world spatial simulation may involve complex physical factors such as object pose, materials and collisions. Since existing open-source models lack such prior knowledge, and to reduce the noise introduced by these additional factors in the evaluation of pure spatial reasoning ability, our task, consistent with prior work on spatial reasoning, does not take these factors into account.
Second, since our benchmark requires arbitrary rotation and translation operations on both objects and agents, collecting such data from real-world environments is both challenging and prohibitively expensive. Therefore, all of our scenes are generated from diverse simulated environments. Nevertheless, our experimental results demonstrate that the learned models generalize well to real-world scenarios. Future work may further explore and improve these two directions.

\section*{Impact Statement}
This paper presents work whose goal is to advance the field of Machine Learning through both a novel methodology and a new evaluation benchmark. Regarding our method, there are many potential societal consequences of our work, none which we feel must be specifically highlighted here. In constructing our benchmark, we ensure strict adherence to copyright and licensing regulations, explicitly avoiding data from sources that prohibit copying or redistribution. Furthermore, we have manually filtered the dataset to exclude any images containing private information or harmful content. The data within our benchmark are not intended to, nor are they capable of, replacing original data sources. Finally, we did not employ external crowdsourcing or paid annotation platforms. All participants contributed voluntarily with a full understanding of the research goals, procedures, and intended data use.

\section*{The Use of Large Language Models}
In this study, a large language model (LLM) was employed as a tool to assist in the refinement and enhancement of the manuscript's language. The specific usages of the LLM include:
\begin{itemize}
    \item \textbf{Grammar and Syntax Improvement:} The LLM helped to correct grammatical errors and improve sentence structures, contributing to greater clarity and fluency in the writing.
    \item \textbf{Conciseness and Precision:} It provided suggestions for more concise and precise wording, aiding in the refinement of certain sections without altering their meaning.
\end{itemize}
It is important to note that while the LLM contributed to the refinement of the manuscript's language, the research ideas, data analysis, and conclusions were independently conceived and developed by the authors. The LLM's contributions were exclusively related to text refinement and did not extend to the conceptual aspects of the study.

\section*{Acknowledgments}
This work was supported by the National Natural Science Foundation of China (No.U24A20335), and the independent research project of the Key Laboratory of Cognition and Decision Intelligence for Complex Systems.

\bibliography{custom}

@inproceedings{mmmu,
  author       = {Xiang Yue and
                  Yuansheng Ni and
                  Tianyu Zheng and
                  Kai Zhang and
                  Ruoqi Liu and
                  Ge Zhang and
                  Samuel Stevens and
                  Dongfu Jiang and
                  Weiming Ren and
                  Yuxuan Sun and
                  Cong Wei and
                  Botao Yu and
                  Ruibin Yuan and
                  Renliang Sun and
                  Ming Yin and
                  Boyuan Zheng and
                  Zhenzhu Yang and
                  Yibo Liu and
                  Wenhao Huang and
                  Huan Sun and
                  Yu Su and
                  Wenhu Chen},
  title        = {{MMMU:} {A} Massive Multi-Discipline Multimodal Understanding and
                  Reasoning Benchmark for Expert {AGI}},
  booktitle    = {{IEEE/CVF} Conference on Computer Vision and Pattern Recognition,
                  {CVPR} 2024, Seattle, WA, USA, June 16-22, 2024},
  pages        = {9556--9567},
  publisher    = {{IEEE}},
  year         = {2024},
  url          = {https://doi.org/10.1109/CVPR52733.2024.00913},
  doi          = {10.1109/CVPR52733.2024.00913},
  bibsource    = {dblp computer science bibliography, https://dblp.org}
}

@inproceedings{thinkinspace,
  author       = {Jihan Yang and
                  Shusheng Yang and
                  Anjali W. Gupta and
                  Rilyn Han and
                  Li Fei{-}Fei and
                  Saining Xie},
  title        = {Thinking in Space: How Multimodal Large Language Models See, Remember,
                  and Recall Spaces},
  booktitle    = {{IEEE/CVF} Conference on Computer Vision and Pattern Recognition,
                  {CVPR} 2025, Nashville, TN, USA, June 11-15, 2025},
  pages        = {10632--10643},
  publisher    = {Computer Vision Foundation / {IEEE}},
  year         = {2025},
  url          = {https://openaccess.thecvf.com/content/CVPR2025/html/Yang\_Thinking\_in\_Space\_How\_Multimodal\_Large\_Language\_Models\_See\_Remember\_CVPR\_2025\_paper.html},
  doi          = {10.1109/CVPR52734.2025.00994},
  bibsource    = {dblp computer science bibliography, https://dblp.org}
}

@article{ViewSpatialbench,
  author       = {Dingming Li and
                  Hongxing Li and
                  Zixuan Wang and
                  Yuchen Yan and
                  Hang Zhang and
                  Siqi Chen and
                  Guiyang Hou and
                  Shengpei Jiang and
                  Wenqiao Zhang and
                  Yongliang Shen and
                  Weiming Lu and
                  Yueting Zhuang},
  title        = {ViewSpatial-Bench: Evaluating Multi-perspective Spatial Localization
                  in Vision-Language Models},
  journal      = {CoRR},
  volume       = {abs/2505.21500},
  year         = {2025},
  url          = {https://doi.org/10.48550/arXiv.2505.21500},
  doi          = {10.48550/ARXIV.2505.21500},
  eprinttype    = {arXiv},
  eprint       = {2505.21500},
  bibsource    = {dblp computer science bibliography, https://dblp.org}
}

@article{mmsibench,
  author       = {Sihan Yang and
                  Runsen Xu and
                  Yiman Xie and
                  Sizhe Yang and
                  Mo Li and
                  Jingli Lin and
                  Chenming Zhu and
                  Xiaochen Chen and
                  Haodong Duan and
                  Xiangyu Yue and
                  Dahua Lin and
                  Tai Wang and
                  Jiangmiao Pang},
  title        = {MMSI-Bench: {A} Benchmark for Multi-Image Spatial Intelligence},
  journal      = {CoRR},
  volume       = {abs/2505.23764},
  year         = {2025},
  url          = {https://doi.org/10.48550/arXiv.2505.23764},
  doi          = {10.48550/ARXIV.2505.23764},
  eprinttype    = {arXiv},
  eprint       = {2505.23764},
  bibsource    = {dblp computer science bibliography, https://dblp.org}
}

@article{qwen2.5-vl,
  author       = {Shuai Bai and
                  Keqin Chen and
                  Xuejing Liu and
                  Jialin Wang and
                  Wenbin Ge and
                  Sibo Song and
                  Kai Dang and
                  Peng Wang and
                  Shijie Wang and
                  Jun Tang and
                  Humen Zhong and
                  Yuanzhi Zhu and
                  Ming{-}Hsuan Yang and
                  Zhaohai Li and
                  Jianqiang Wan and
                  Pengfei Wang and
                  Wei Ding and
                  Zheren Fu and
                  Yiheng Xu and
                  Jiabo Ye and
                  Xi Zhang and
                  Tianbao Xie and
                  Zesen Cheng and
                  Hang Zhang and
                  Zhibo Yang and
                  Haiyang Xu and
                  Junyang Lin},
  title        = {Qwen2.5-VL Technical Report},
  journal      = {CoRR},
  volume       = {abs/2502.13923},
  year         = {2025},
  url          = {https://doi.org/10.48550/arXiv.2502.13923},
  doi          = {10.48550/ARXIV.2502.13923},
  eprinttype    = {arXiv},
  eprint       = {2502.13923},
  bibsource    = {dblp computer science bibliography, https://dblp.org}
}

@article{gemma3,
  author       = {Aishwarya Kamath and
                  Johan Ferret and
                  Shreya Pathak and
                  Nino Vieillard and
                  Ramona Merhej and
                  Sarah Perrin and
                  Tatiana Matejovicova and
                  Alexandre Ram{\'{e}} and
                  Morgane Rivi{\`{e}}re and
                  Louis Rouillard and
                  Thomas Mesnard and
                  Geoffrey Cideron and
                  Jean{-}Bastien Grill and
                  Sabela Ramos and
                  Edouard Yvinec and
                  Michelle Casbon and
                  Etienne Pot and
                  Ivo Penchev and
                  Ga{\"{e}}l Liu and
                  Francesco Visin and
                  Kathleen Kenealy and
                  Lucas Beyer and
                  Xiaohai Zhai and
                  Anton Tsitsulin and
                  R{\'{o}}bert Busa{-}Fekete and
                  Alex Feng and
                  Noveen Sachdeva and
                  Benjamin Coleman and
                  Yi Gao and
                  Basil Mustafa and
                  Iain Barr and
                  Emilio Parisotto and
                  David Tian and
                  Matan Eyal and
                  Colin Cherry and
                  Jan{-}Thorsten Peter and
                  Danila Sinopalnikov and
                  Surya Bhupatiraju and
                  Rishabh Agarwal and
                  Mehran Kazemi and
                  Dan Malkin and
                  Ravin Kumar and
                  David Vilar and
                  Idan Brusilovsky and
                  Jiaming Luo and
                  Andreas Steiner and
                  Abe Friesen and
                  Abhanshu Sharma and
                  Abheesht Sharma and
                  Adi Mayrav Gilady and
                  Adrian Goedeckemeyer and
                  Alaa Saade and
                  Alexander Kolesnikov and
                  Alexei Bendebury and
                  Alvin Abdagic and
                  Amit Vadi and
                  Andr{\'{a}}s Gy{\"{o}}rgy and
                  Andr{\'{e}} Susano Pinto and
                  Anil Das and
                  Ankur Bapna and
                  Antoine Miech and
                  Antoine Yang and
                  Antonia Paterson and
                  Ashish Shenoy and
                  Ayan Chakrabarti and
                  Bilal Piot and
                  Bo Wu and
                  Bobak Shahriari and
                  Bryce Petrini and
                  Charlie Chen and
                  Charline Le Lan and
                  Christopher A. Choquette{-}Choo and
                  CJ Carey and
                  Cormac Brick and
                  Daniel Deutsch and
                  Danielle Eisenbud and
                  Dee Cattle and
                  Derek Cheng and
                  Dimitris Paparas and
                  Divyashree Shivakumar Sreepathihalli and
                  Doug Reid and
                  Dustin Tran and
                  Dustin Zelle and
                  Eric Noland and
                  Erwin Huizenga and
                  Eugene Kharitonov and
                  Frederick Liu and
                  Gagik Amirkhanyan and
                  Glenn Cameron and
                  Hadi Hashemi and
                  Hanna Klimczak{-}Plucinska and
                  Harman Singh and
                  Harsh Mehta and
                  Harshal Tushar Lehri and
                  Hussein Hazimeh and
                  Ian Ballantyne and
                  Idan Szpektor and
                  Ivan Nardini and
                  Jean Pouget{-}Abadie and
                  Jetha Chan and
                  Joe Stanton and
                  John Wieting and
                  Jonathan Lai and
                  Jordi Orbay and
                  Joseph Fernandez and
                  Josh Newlan and
                  Ju{-}yeong Ji and
                  Jyotinder Singh and
                  Kat Black and
                  Kathy Yu and
                  Kevin Hui and
                  Kiran Vodrahalli and
                  Klaus Greff and
                  Linhai Qiu and
                  Marcella Valentine and
                  Marina Coelho and
                  Marvin Ritter and
                  Matt Hoffman and
                  Matthew Watson and
                  Mayank Chaturvedi and
                  Michael Moynihan and
                  Min Ma and
                  Nabila Babar and
                  Natasha Noy and
                  Nathan Byrd and
                  Nick Roy and
                  Nikola Momchev and
                  Nilay Chauhan and
                  Oskar Bunyan and
                  Pankil Botarda and
                  Paul Caron and
                  Paul Kishan Rubenstein and
                  Phil Culliton and
                  Philipp Schmid and
                  Pier Giuseppe Sessa and
                  Pingmei Xu and
                  Piotr Stanczyk and
                  Pouya Tafti and
                  Rakesh Shivanna and
                  Renjie Wu and
                  Renke Pan and
                  Reza Rokni and
                  Rob Willoughby and
                  Rohith Vallu and
                  Ryan Mullins and
                  Sammy Jerome and
                  Sara Smoot and
                  Sertan Girgin and
                  Shariq Iqbal and
                  Shashir Reddy and
                  Shruti Sheth and
                  Siim P{\~{o}}der and
                  Sijal Bhatnagar and
                  Sindhu Raghuram Panyam and
                  Sivan Eiger and
                  Susan Zhang and
                  Tianqi Liu and
                  Trevor Yacovone and
                  Tyler Liechty and
                  Uday Kalra and
                  Utku Evci and
                  Vedant Misra and
                  Vincent Roseberry and
                  Vlad Feinberg and
                  Vlad Kolesnikov and
                  Woohyun Han and
                  Woosuk Kwon and
                  Xi Chen and
                  Yinlam Chow and
                  Yuvein Zhu and
                  Zichuan Wei and
                  Zoltan Egyed and
                  Victor Cotruta and
                  Minh Giang and
                  Phoebe Kirk and
                  Anand Rao and
                  Jessica Lo and
                  Erica Moreira and
                  Luiz Gustavo Martins and
                  Omar Sanseviero and
                  Lucas Gonzalez and
                  Zach Gleicher and
                  Tris Warkentin and
                  Vahab Mirrokni and
                  Evan Senter and
                  Eli Collins and
                  Joelle K. Barral and
                  Zoubin Ghahramani and
                  Raia Hadsell and
                  Yossi Matias and
                  D. Sculley and
                  Slav Petrov and
                  Noah Fiedel and
                  Noam Shazeer and
                  Oriol Vinyals and
                  Jeff Dean and
                  Demis Hassabis and
                  Koray Kavukcuoglu and
                  Cl{\'{e}}ment Farabet and
                  Elena Buchatskaya and
                  Jean{-}Baptiste Alayrac and
                  Rohan Anil and
                  Dmitry (Dima) Lepikhin and
                  Sebastian Borgeaud and
                  Olivier Bachem and
                  Armand Joulin and
                  Alek Andreev and
                  Cassidy Hardin and
                  Robert Dadashi and
                  L{\'{e}}onard Hussenot},
  title        = {Gemma 3 Technical Report},
  journal      = {CoRR},
  volume       = {abs/2503.19786},
  year         = {2025},
  url          = {https://doi.org/10.48550/arXiv.2503.19786},
  doi          = {10.48550/ARXIV.2503.19786},
  eprinttype    = {arXiv},
  eprint       = {2503.19786},
  bibsource    = {dblp computer science bibliography, https://dblp.org}
}

@misc{internvl-3.5,
      title={InternVL3.5: Advancing Open-Source Multimodal Models in Versatility, Reasoning, and Efficiency}, 
      author={Weiyun Wang and Zhangwei Gao and Lixin Gu and Hengjun Pu and Long Cui and Xingguang Wei and Zhaoyang Liu and Linglin Jing and Shenglong Ye and Jie Shao and Zhaokai Wang and Zhe Chen and Hongjie Zhang and Ganlin Yang and Haomin Wang and Qi Wei and Jinhui Yin and Wenhao Li and Erfei Cui and Guanzhou Chen and Zichen Ding and Changyao Tian and Zhenyu Wu and Jingjing Xie and Zehao Li and Bowen Yang and Yuchen Duan and Xuehui Wang and Zhi Hou and Haoran Hao and Tianyi Zhang and Songze Li and Xiangyu Zhao and Haodong Duan and Nianchen Deng and Bin Fu and Yinan He and Yi Wang and Conghui He and Botian Shi and Junjun He and Yingtong Xiong and Han Lv and Lijun Wu and Wenqi Shao and Kaipeng Zhang and Huipeng Deng and Biqing Qi and Jiaye Ge and Qipeng Guo and Wenwei Zhang and Songyang Zhang and Maosong Cao and Junyao Lin and Kexian Tang and Jianfei Gao and Haian Huang and Yuzhe Gu and Chengqi Lyu and Huanze Tang and Rui Wang and Haijun Lv and Wanli Ouyang and Limin Wang and Min Dou and Xizhou Zhu and Tong Lu and Dahua Lin and Jifeng Dai and Weijie Su and Bowen Zhou and Kai Chen and Yu Qiao and Wenhai Wang and Gen Luo},
      year={2025},
      eprint={2508.18265},
      archivePrefix={arXiv},
      primaryClass={cs.CV},
      url={https://arxiv.org/abs/2508.18265}, 
}

@article{gridcell1,
  title={Grid-cell representations in mental simulation},
  author={Bellmund, Jacob LS and Deuker, Lorena and Navarro Schr{\"o}der, Tobias and Doeller, Christian F},
  journal={Elife},
  volume={5},
  pages={e17089},
  year={2016},
  publisher={eLife Sciences Publications, Ltd}
}

@article{gridcell2,
  title={Spatial representations in the human brain},
  author={Herweg, Nora A and Kahana, Michael J},
  journal={Frontiers in human neuroscience},
  volume={12},
  pages={297},
  year={2018},
  publisher={Frontiers Media SA}
}

@article{gridcell3,
  title={Distance and grid-like codes support the navigation of abstract social space in the human brain},
  author={Liang, Zilu and Wu, Simeng and Wu, Jie and Wang, Wen-Xu and Qin, Shaozheng and Liu, Chao},
  journal={Elife},
  volume={12},
  pages={RP89025},
  year={2024},
  publisher={eLife Sciences Publications Limited}
}

@article{spatialladder,
  title={Spatialladder: Progressive training for spatial reasoning in vision-language models},
  author={Li, Hongxing and Li, Dingming and Wang, Zixuan and Yan, Yuchen and Wu, Hang and Zhang, Wenqi and Shen, Yongliang and Lu, Weiming and Xiao, Jun and Zhuang, Yueting},
  journal={arXiv preprint arXiv:2510.08531},
  year={2025}
}

@article{spatialthinker,
  title={SpatialThinker: Reinforcing 3D Reasoning in Multimodal LLMs via Spatial Rewards},
  author={Batra, Hunar and Tu, Haoqin and Chen, Hardy and Lin, Yuanze and Xie, Cihang and Clark, Ronald},
  journal={arXiv preprint arXiv:2511.07403},
  year={2025}
}

@article{spatialssrl,
  title={Spatial-ssrl: Enhancing spatial understanding via self-supervised reinforcement learning},
  author={Liu, Yuhong and Zhang, Beichen and Zang, Yuhang and Cao, Yuhang and Xing, Long and Dong, Xiaoyi and Duan, Haodong and Lin, Dahua and Wang, Jiaqi},
  journal={arXiv preprint arXiv:2510.27606},
  year={2025}
}

@article{vilasr,
  title={Reinforcing spatial reasoning in vision-language models with interwoven thinking and visual drawing},
  author={Wu, Junfei and Guan, Jian and Feng, Kaituo and Liu, Qiang and Wu, Shu and Wang, Liang and Wu, Wei and Tan, Tieniu},
  journal={arXiv preprint arXiv:2506.09965},
  year={2025}
}

@inproceedings{habitat,
  title={Habitat: A platform for embodied ai research},
  author={Savva, Manolis and Kadian, Abhishek and Maksymets, Oleksandr and Zhao, Yili and Wijmans, Erik and Jain, Bhavana and Straub, Julian and Liu, Jia and Koltun, Vladlen and Malik, Jitendra and others},
  booktitle={Proceedings of the IEEE/CVF international conference on computer vision},
  pages={9339--9347},
  year={2019}
}

@article{replica,
  title={The replica dataset: A digital replica of indoor spaces},
  author={Straub, Julian and Whelan, Thomas and Ma, Lingni and Chen, Yufan and Wijmans, Erik and Green, Simon and Engel, Jakob J and Mur-Artal, Raul and Ren, Carl and Verma, Shobhit and others},
  journal={arXiv preprint arXiv:1906.05797},
  year={2019}
}

@article{ai2thor,
  title={Ai2-thor: An interactive 3d environment for visual ai},
  author={Kolve, Eric and Mottaghi, Roozbeh and Han, Winson and VanderBilt, Eli and Weihs, Luca and Herrasti, Alvaro and Deitke, Matt and Ehsani, Kiana and Gordon, Daniel and Zhu, Yuke and others},
  journal={arXiv preprint arXiv:1712.05474},
  year={2017}
}

@article{spatialtree,
  title={SpatialTree: How Spatial Abilities Branch Out in MLLMs},
  author={Xiao, Yuxi and Li, Longfei and Yan, Shen and Liu, Xinhang and Peng, Sida and Wei, Yunchao and Zhou, Xiaowei and Kang, Bingyi},
  journal={arXiv preprint arXiv:2512.20617},
  year={2025}
}

@inproceedings{spatialcontext,
  title={Spatial reasoning in context: bridging cognitive and educational perspectives of spatial-mathematics relations},
  author={Harris, Danielle},
  booktitle={Frontiers in Education},
  volume={8},
  pages={1302099},
  year={2023},
  organization={Frontiers Media SA}
}

@article{density,
  title={Distance to nearest neighbor as a measure of spatial relationships in populations},
  author={Clark, Philip J and Evans, Francis C},
  journal={Ecology},
  volume={35},
  number={4},
  pages={445--453},
  year={1954},
  publisher={JSTOR}
}

@article{stare,
  author       = {Linjie Li and
                  Mahtab Bigverdi and
                  Jiawei Gu and
                  Zixian Ma and
                  Yinuo Yang and
                  Ziang Li and
                  Yejin Choi and
                  Ranjay Krishna},
  title        = {Unfolding Spatial Cognition: Evaluating Multimodal Models on Visual
                  Simulations},
  journal      = {CoRR},
  volume       = {abs/2506.04633},
  year         = {2025},
  url          = {https://doi.org/10.48550/arXiv.2506.04633},
  doi          = {10.48550/ARXIV.2506.04633},
  eprinttype    = {arXiv},
  eprint       = {2506.04633},
  bibsource    = {dblp computer science bibliography, https://dblp.org}
}

@article{mindcube,
  author       = {Baiqiao Yin and
                  Qineng Wang and
                  Pingyue Zhang and
                  Jianshu Zhang and
                  Kangrui Wang and
                  Zihan Wang and
                  Jieyu Zhang and
                  Keshigeyan Chandrasegaran and
                  Han Liu and
                  Ranjay Krishna and
                  Saining Xie and
                  Manling Li and
                  Jiajun Wu and
                  Li Fei{-}Fei},
  title        = {Spatial Mental Modeling from Limited Views},
  journal      = {CoRR},
  volume       = {abs/2506.21458},
  year         = {2025},
  url          = {https://doi.org/10.48550/arXiv.2506.21458},
  doi          = {10.48550/ARXIV.2506.21458},
  eprinttype    = {arXiv},
  eprint       = {2506.21458},
  bibsource    = {dblp computer science bibliography, https://dblp.org}
}

@article{sims-v,
  author       = {Ellis Brown and
                  Arijit Ray and
                  Ranjay Krishna and
                  Ross B. Girshick and
                  Rob Fergus and
                  Saining Xie},
  title        = {{SIMS-V:} Simulated Instruction-Tuning for Spatial Video Understanding},
  journal      = {CoRR},
  volume       = {abs/2511.04668},
  year         = {2025},
  url          = {https://doi.org/10.48550/arXiv.2511.04668},
  doi          = {10.48550/ARXIV.2511.04668},
  eprinttype    = {arXiv},
  eprint       = {2511.04668},
  bibsource    = {dblp computer science bibliography, https://dblp.org}
}

@article{spatial_survey,
  author       = {Weichen Liu and
                  Qiyao Xue and
                  Haoming Wang and
                  Xiangyu Yin and
                  Boyuan Yang and
                  Wei Gao},
  title        = {Spatial Reasoning in Multimodal Large Language Models: {A} Survey
                  of Tasks, Benchmarks and Methods},
  journal      = {CoRR},
  volume       = {abs/2511.15722},
  year         = {2025},
  url          = {https://doi.org/10.48550/arXiv.2511.15722},
  doi          = {10.48550/ARXIV.2511.15722},
  eprinttype    = {arXiv},
  eprint       = {2511.15722},
  bibsource    = {dblp computer science bibliography, https://dblp.org}
}

@article{spatial-mllm,
  author       = {Diankun Wu and
                  Fangfu Liu and
                  Yi{-}Hsin Hung and
                  Yueqi Duan},
  title        = {Spatial-MLLM: Boosting {MLLM} Capabilities in Visual-based Spatial
                  Intelligence},
  journal      = {CoRR},
  volume       = {abs/2505.23747},
  year         = {2025},
  url          = {https://doi.org/10.48550/arXiv.2505.23747},
  doi          = {10.48550/ARXIV.2505.23747},
  eprinttype    = {arXiv},
  eprint       = {2505.23747},
  bibsource    = {dblp computer science bibliography, https://dblp.org}
}

@article{embodied-r,
  author       = {Baining Zhao and
                  Ziyou Wang and
                  Jianjie Fang and
                  Chen Gao and
                  Fanhang Man and
                  Jinqiang Cui and
                  Xin Wang and
                  Xinlei Chen and
                  Yong Li and
                  Wenwu Zhu},
  title        = {Embodied-R: Collaborative Framework for Activating Embodied Spatial
                  Reasoning in Foundation Models via Reinforcement Learning},
  journal      = {CoRR},
  volume       = {abs/2504.12680},
  year         = {2025},
  url          = {https://doi.org/10.48550/arXiv.2504.12680},
  doi          = {10.48550/ARXIV.2504.12680},
  eprinttype    = {arXiv},
  eprint       = {2504.12680},
  bibsource    = {dblp computer science bibliography, https://dblp.org}
}

@article{visualpuzzle,
  author       = {Yueqi Song and
                  Tianyue Ou and
                  Yibo Kong and
                  Zecheng Li and
                  Graham Neubig and
                  Xiang Yue},
  title        = {VisualPuzzles: Decoupling Multimodal Reasoning Evaluation from Domain
                  Knowledge},
  journal      = {CoRR},
  volume       = {abs/2504.10342},
  year         = {2025},
  url          = {https://doi.org/10.48550/arXiv.2504.10342},
  doi          = {10.48550/ARXIV.2504.10342},
  eprinttype    = {arXiv},
  eprint       = {2504.10342},
  bibsource    = {dblp computer science bibliography, https://dblp.org}
}

@article{ominispatial,
  author       = {Mengdi Jia and
                  Zekun Qi and
                  Shaochen Zhang and
                  Wenyao Zhang and
                  Xinqiang Yu and
                  Jiawei He and
                  He Wang and
                  Li Yi},
  title        = {OmniSpatial: Towards Comprehensive Spatial Reasoning Benchmark for
                  Vision Language Models},
  journal      = {CoRR},
  volume       = {abs/2506.03135},
  year         = {2025},
  url          = {https://doi.org/10.48550/arXiv.2506.03135},
  doi          = {10.48550/ARXIV.2506.03135},
  eprinttype    = {arXiv},
  eprint       = {2506.03135},
  bibsource    = {dblp computer science bibliography, https://dblp.org}
}

@inproceedings{vllm,
  title={Efficient Memory Management for Large Language Model Serving with PagedAttention},
  author={Woosuk Kwon and Zhuohan Li and Siyuan Zhuang and Ying Sheng and Lianmin Zheng and Cody Hao Yu and Joseph E. Gonzalez and Hao Zhang and Ion Stoica},
  booktitle={Proceedings of the ACM SIGOPS 29th Symposium on Operating Systems Principles},
  year={2023}
}

@article{grpo,
  author       = {Zhihong Shao and
                  Peiyi Wang and
                  Qihao Zhu and
                  Runxin Xu and
                  Junxiao Song and
                  Mingchuan Zhang and
                  Y. K. Li and
                  Y. Wu and
                  Daya Guo},
  title        = {DeepSeekMath: Pushing the Limits of Mathematical Reasoning in Open
                  Language Models},
  journal      = {CoRR},
  volume       = {abs/2402.03300},
  year         = {2024},
  url          = {https://doi.org/10.48550/arXiv.2402.03300},
  doi          = {10.48550/ARXIV.2402.03300},
  eprinttype    = {arXiv},
  eprint       = {2402.03300},
  bibsource    = {dblp computer science bibliography, https://dblp.org}
}

@article{mmr-v,
  author       = {Kejian Zhu and
                  Zhuoran Jin and
                  Hongbang Yuan and
                  Jiachun Li and
                  Shangqing Tu and
                  Pengfei Cao and
                  Yubo Chen and
                  Kang Liu and
                  Jun Zhao},
  title        = {{MMR-V:} What's Left Unsaid? {A} Benchmark for Multimodal Deep
                  Reasoning in Videos},
  journal      = {CoRR},
  volume       = {abs/2506.04141},
  year         = {2025},
  url          = {https://doi.org/10.48550/arXiv.2506.04141},
  doi          = {10.48550/ARXIV.2506.04141},
  eprinttype    = {arXiv},
  eprint       = {2506.04141},
  bibsource    = {dblp computer science bibliography, https://dblp.org}
}

@online{gemini3,
  author = {{Google DeepMind}},
  title = {Gemini 3},
  year = {2025},
  url = {https://deepmind.google/technologies/gemini/},
  note = {Accessed: 2026-01-29}
}

@online{gpt-5,
  author    = {{OpenAI}},
  title     = {Introducing GPT-5},
  year      = {2025},
  url       = {https://openai.com/index/introducing-gpt-5/},
  note      = {Accessed: 2026-01-29}
}

@online{gpt-4o,
  author    = {{OpenAI}},
  title     = {Hello GPT-4o},
  year      = {2024},
  url       = {https://openai.com/index/hello-gpt-4o/},
  note      = {Accessed: 2026-01-29}
}

@inproceedings{apc,
  title={Perspective-aware reasoning in vision-language models via mental imagery simulation},
  author={Lee, Phillip Y and Je, Jihyeon and Park, Chanho and Uy, Mikaela Angelina and Guibas, Leonidas and Sung, Minhyuk},
  booktitle={Proceedings of the IEEE/CVF International Conference on Computer Vision (ICCV)},
  year={2025}
}

@article{spacer,
  title={SpaceR: Reinforcing MLLMs in Video Spatial Reasoning},
  author={Ouyang, Kun and Liu, Yuanxin and Wu, Haoning and Liu, Yi and Zhou, Hao and Zhou, Jie and Meng, Fandong and Sun, Xu},
  journal={arXiv preprint arXiv:2504.01805},
  year={2025}
}

@inproceedings{spatialeval,
        title={Is A Picture Worth A Thousand Words? Delving Into Spatial Reasoning for Vision Language Models},
        author={Wang, Jiayu and Ming, Yifei and Shi, Zhenmei and Vineet, Vibhav and Wang, Xin and Li, Yixuan and Joshi, Neel},
        booktitle={The Thirty-Eighth Annual Conference on Neural Information Processing Systems},
        year={2024}
      }

@article{spatialviz,
  title={Spatialviz-bench: An mllm benchmark for spatial visualization},
  author={Wang, Siting and Pei, Minnan and Sun, Luoyang and Deng, Cheng and Shao, Kun and Tian, Zheng and Zhang, Haifeng and Wang, Jun},
  journal={arXiv preprint arXiv:2507.07610},
  year={2025}
}

@article{MulSeT,
  author       = {Wanyue Zhang and
                  Yibin Huang and
                  Yangbin Xu and
                  JingJing Huang and
                  Helu Zhi and
                  Shuo Ren and
                  Wang Xu and
                  Jiajun Zhang},
  title        = {Why Do MLLMs Struggle with Spatial Understanding? {A} Systematic Analysis
                  from Data to Architecture},
  journal      = {CoRR},
  volume       = {abs/2509.02359},
  year         = {2025},
  url          = {https://doi.org/10.48550/arXiv.2509.02359},
  doi          = {10.48550/ARXIV.2509.02359},
  eprinttype   = {arXiv},
  eprint       = {2509.02359},
  bibsource    = {dblp computer science bibliography, https://dblp.org}
}

@article{SpinBench,
  author       = {Yuyou Zhang and
                  Radu Corcodel and
                  Chiori Hori and
                  Anoop Cherian and
                  Ding Zhao},
  title        = {SpinBench: Perspective and Rotation as a Lens on Spatial Reasoning
                  in VLMs},
  journal      = {CoRR},
  volume       = {abs/2509.25390},
  year         = {2025},
  url          = {https://doi.org/10.48550/arXiv.2509.25390},
  doi          = {10.48550/ARXIV.2509.25390},
  eprinttype   = {arXiv},
  eprint       = {2509.25390},
  bibsource    = {dblp computer science bibliography, https://dblp.org}
}

@article{MindJourney,
  author       = {Yuncong Yang and
                  Jiageng Liu and
                  Zheyuan Zhang and
                  Siyuan Zhou and
                  Reuben Tan and
                  Jianwei Yang and
                  Yilun Du and
                  Chuang Gan},
  title        = {MindJourney: Test-Time Scaling with World Models for Spatial Reasoning},
  journal      = {CoRR},
  volume       = {abs/2507.12508},
  year         = {2025},
  url          = {https://doi.org/10.48550/arXiv.2507.12508},
  doi          = {10.48550/ARXIV.2507.12508},
  eprinttype   = {arXiv},
  eprint       = {2507.12508},
  bibsource    = {dblp computer science bibliography, https://dblp.org}
}

@article{qwen3-vl,
  author       = {Qwen Team},
  title        = {Qwen3-VL Technical Report},
  journal      = {CoRR},
  volume       = {abs/2511.21631},
  year         = {2025},
  url          = {https://doi.org/10.48550/arXiv.2511.21631},
  doi          = {10.48550/ARXIV.2511.21631},
  eprinttype   = {arXiv},
  eprint       = {2511.21631},
  bibsource    = {dblp computer science bibliography, https://dblp.org}
}

@inproceedings{llamafactory,
  title={LlamaFactory: Unified Efficient Fine-Tuning of 100+ Language Models},
  author={Yaowei Zheng and Richong Zhang and Junhao Zhang and Yanhan Ye and Zheyan Luo and Zhangchi Feng and Yongqiang Ma},
  booktitle={Proceedings of the 62nd Annual Meeting of the Association for Computational Linguistics (Volume 3: System Demonstrations)},
  address={Bangkok, Thailand},
  publisher={Association for Computational Linguistics},
  year={2024},
  url={http://arxiv.org/abs/2403.13372}
}

@article{verl,
  title   = {HybridFlow: A Flexible and Efficient RLHF Framework},
  author  = {Guangming Sheng and Chi Zhang and Zilingfeng Ye and Xibin Wu and Wang Zhang and Ru Zhang and Yanghua Peng and Haibin Lin and Chuan Wu},
  year    = {2024},
  journal = {arXiv preprint arXiv: 2409.19256}
}

@online{gemini3.1,
  author = {{Google DeepMind}},
  title = {Gemini 3.1 Pro},
  year = {2025},
  url = {https://deepmind.google/models/gemini/pro/},
  note = {Accessed: 2026-03-16}
}

@inproceedings{scanet,
  author       = {Angela Dai and
                  Angel X. Chang and
                  Manolis Savva and
                  Maciej Halber and
                  Thomas A. Funkhouser and
                  Matthias Nie{\ss}ner},
  title        = {ScanNet: Richly-Annotated 3D Reconstructions of Indoor Scenes},
  booktitle    = {2017 {IEEE} Conference on Computer Vision and Pattern Recognition,
                  {CVPR} 2017, Honolulu, HI, USA, July 21-26, 2017},
  pages        = {2432--2443},
  publisher    = {{IEEE} Computer Society},
  year         = {2017},
  url          = {https://doi.org/10.1109/CVPR.2017.261},
  doi          = {10.1109/CVPR.2017.261},
  bibsource    = {dblp computer science bibliography, https://dblp.org}
}

@misc{arkit,
      title={ARKitScenes: A Diverse Real-World Dataset For 3D Indoor Scene Understanding Using Mobile RGB-D Data}, 
      author={Gilad Baruch and Zhuoyuan Chen and Afshin Dehghan and Tal Dimry and Yuri Feigin and Peter Fu and Thomas Gebauer and Brandon Joffe and Daniel Kurz and Arik Schwartz and Elad Shulman},
      year={2022},
      eprint={2111.08897},
      archivePrefix={arXiv},
      primaryClass={cs.CV},
      url={https://arxiv.org/abs/2111.08897}, 
}
\clearpage
\appendix

\section{Details of Our MentalGrid Benchmark} \label{append:benchmark}

\subsection{Dataset Statistics}
Table \ref{tab:stats} reports the key statistics of our MentalGrid benchmark. We calculate the average response length using the correct answers from GPT-5-mini.

\begin{table}[ht]
\centering

\begin{tabular}{@{}l r@{}}
\toprule
\textbf{Statistics} & \textbf{Number} \\
\midrule
Total Questions & 3,356 \\
Total Query Types & 24 \\
Total Images & 1,100 \\
Total Secnes & 349 \\
\midrule
Task Type \\ 
\quad - Static Spatial Perception & 1162 (34.62\%) \\
\quad - Final State Prediction  & 1143 (34.06\%) \\
\quad - Dynamic Process Tracking & 1051 (31.32\%) \\
\midrule
Multi-choice Questions & 2310 (68.83\%) \\
Open-Ended Questions & 1046 (31.17\%) \\
\midrule
Average Question Length & 449.95 \\
Average Response Length & 1156.04 \\
\bottomrule
\end{tabular}
\caption{Key statistics of MentalGrid.}\label{tab:stats}
\end{table}

\subsection{Comparison with Other Benchmarks}
To further distinguish MentalGrid from existing spatial reasoning benchmarks, we provide a detailed comparison in Table \ref{tab:comp}, highlighting its advantages in three key aspects: \textbf{(1) Data Integrity and Novelty:} Unlike many prior datasets that synthesize data from common sources like ScanNet \citep{scanet} or ARKitScenes \citep{arkit}, which carry a high risk of data contamination, our dataset generates scenes and objects within a custom simulation environment, ensuring spatial novelty and mitigating performance inflation. \textbf{(2) Dynamic Agent-Environment Interaction:} While current benchmarks are primarily restricted to limited viewpoint switching within static environments, MentalGrid features arbitrary agent perspectives and dynamic environmental state changes, necessitating that models move beyond static spatial perception to actively track evolving states. \textbf{(3) Long-Horizon Simulation Complexity:} Our simulation trajectories are designed with significantly greater depth, demanding sophisticated long-horizon spatial simulation capabilities to resolve complex tasks. 

\begin{figure*}[h]
    \centering
    \includegraphics[width=0.9\linewidth]{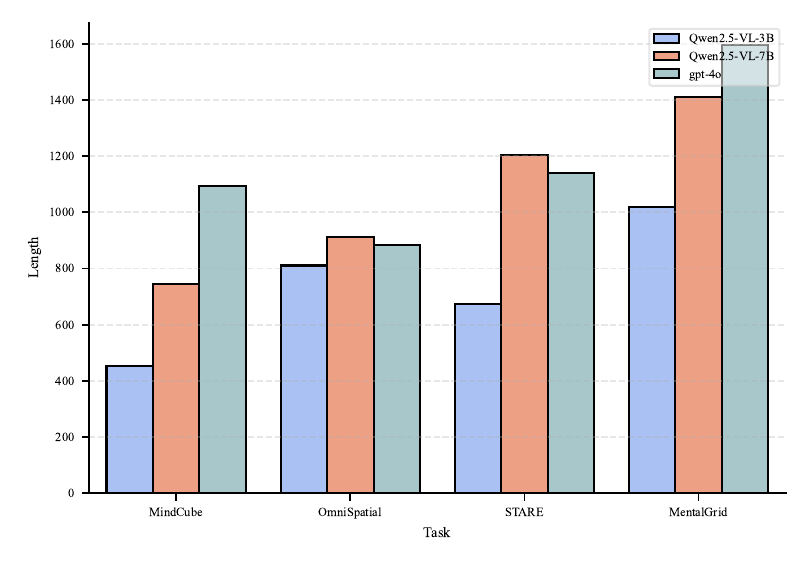}
    \caption{Response length comparison on different benchmarks.} \label{fig:response_length}
\end{figure*}

To further demonstrate the simulation complexity of our tasks, we compare the average reasoning path lengths of various models across MentalGrid and other existing mental simulation datasets in Figure \ref{fig:response_length}. The results indicate that models consistently generate significantly longer trajectories on our dataset, which directly validates the higher simulation complexity of MentalGrid.

\begin{table*}[h]
\centering

\resizebox{\linewidth}{!}{
\begin{tabular}{lccccccc}
\toprule
\textbf{Dataset} & \textbf{Size} & \textbf{Source} & \textbf{Image} & \textbf{View}  & \textbf{Env.}  & \textbf{Sim. Complexity} \\ 
\midrule
SpatialEval \citep{spatialeval}  & 4.6K  & S+E & Mixed & Static & Static & None \\
VSI-Bench \citep{thinkinspace} & 5K & E & Real-life & Fixed & Static & Low \\
ViewSpatial-Bench \citep{ViewSpatialbench} & 5.7K & E & Real-life  & Fixed & Static & Low\\ 
MMSI-Bench \citep{mmsibench}& 1K & E & Real-life &  Fixed  & Static & Low   \\ 
MindCube \citep{mindcube} & 21.1K & E & Real-life  &  Fixed & Static & Medium \\
OminiSpatial \citep{ominispatial} & 8.4K & E & Mixed & Fixed & Static & Medium \\ 
\midrule
\textbf{MentalGrid (Ours)} & \textbf{3.5K} & \textbf{S} & \textbf{Simulation}    & \textbf{Arbitrary}    &  \textbf{Dynamic}    & \textbf{High}  \\ 
\bottomrule
\end{tabular}}
\caption{The comparison between MentalGrid and other existing benchmarks. For Data Source, S denotes synthetic data, while E represents data derived from existing sources. View and Env. indicate whether the agent's viewpoint and the environment state are static or dynamic. Sim. Complexity assesses the overall difficulty of the mental simulation required for each task.} \label{tab:comp}

\end{table*}

\subsection{More Examples in MentalGrid}
\begin{figure*}[h]
    \centering
    \includegraphics[width=0.9\linewidth]{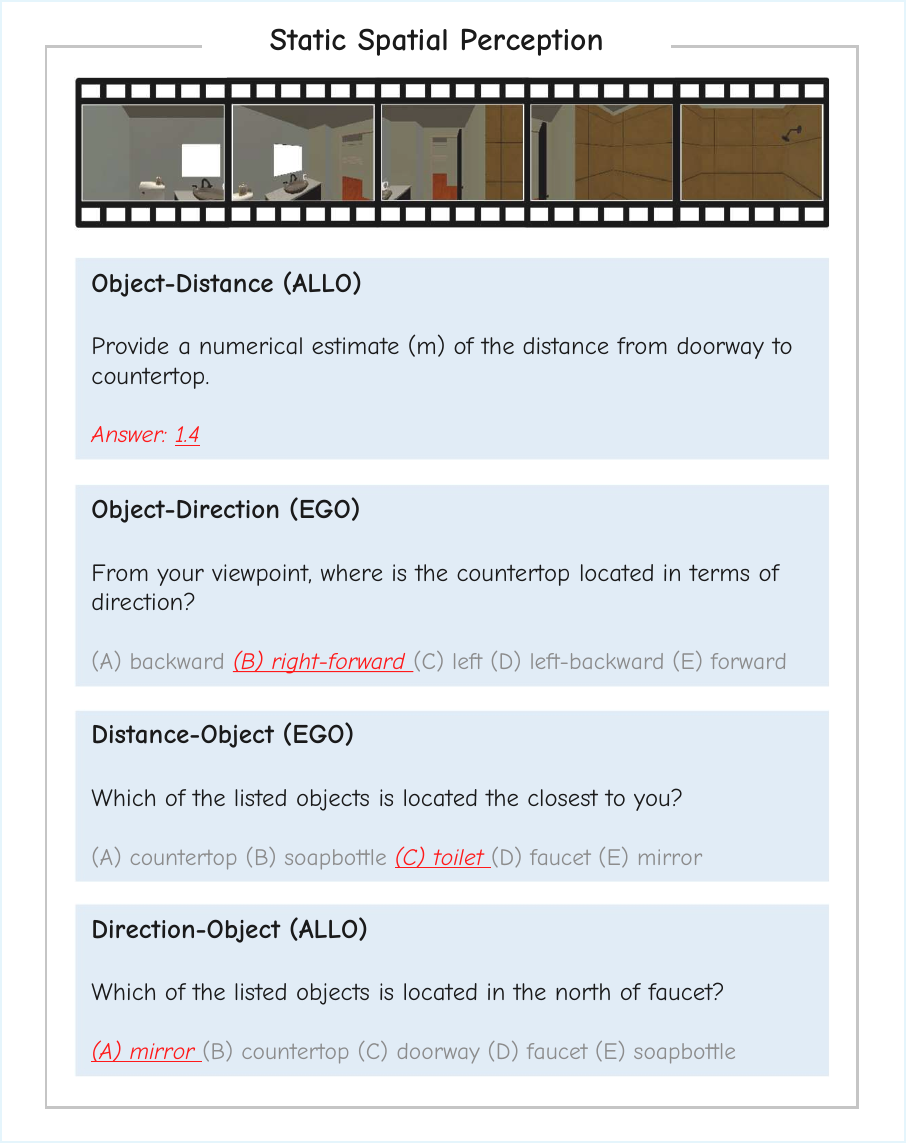}
    \caption{Examples in MentalGrid (Part 1).} \label{fig:case1}
\end{figure*}

\begin{figure*}[h]
    \centering
    \includegraphics[width=0.7\linewidth]{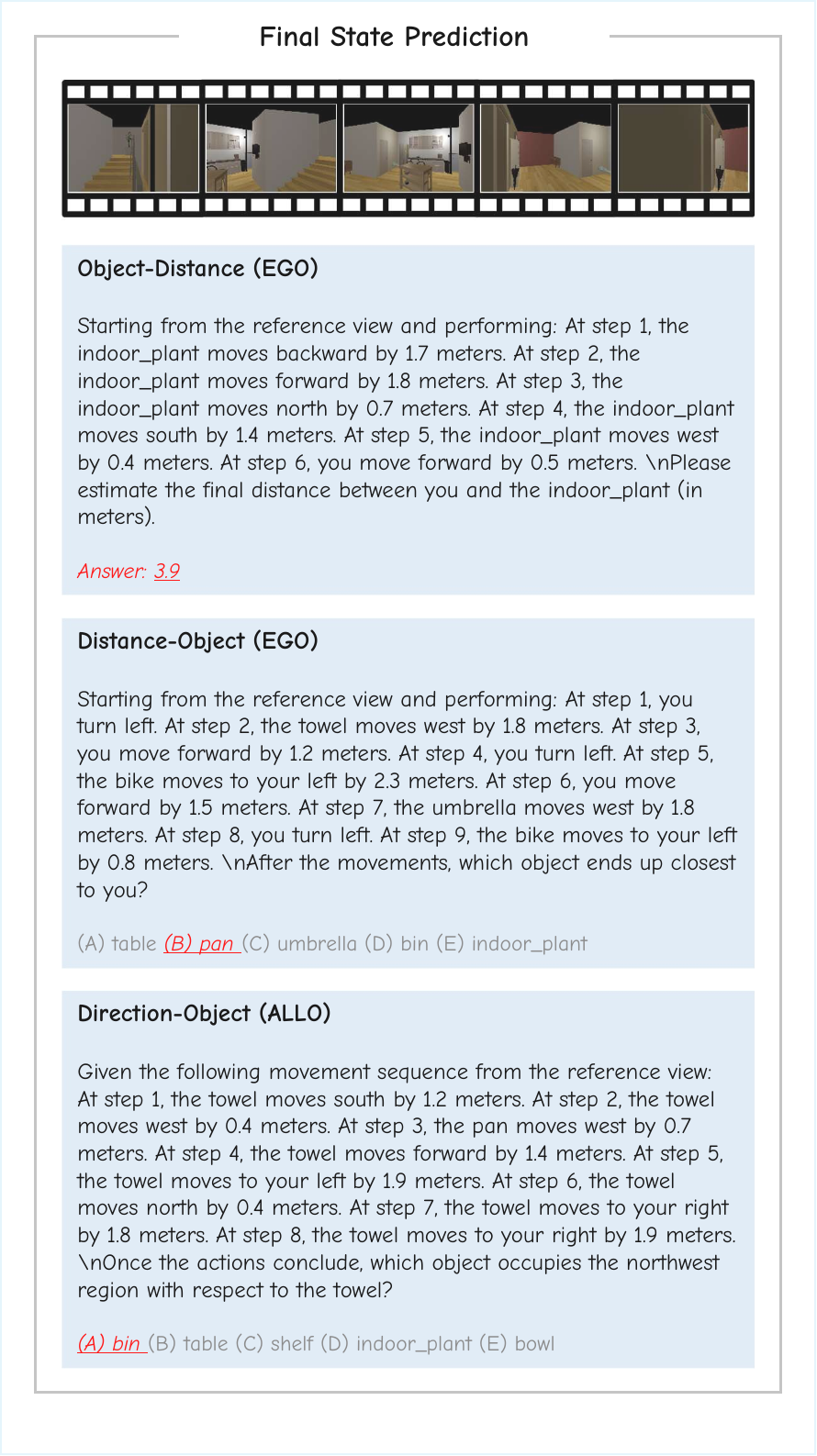}
    \caption{Examples in MentalGrid (Part 2).} \label{fig:case2}
\end{figure*}

\begin{figure*}[h]
    \centering
    \includegraphics[width=0.8\linewidth]{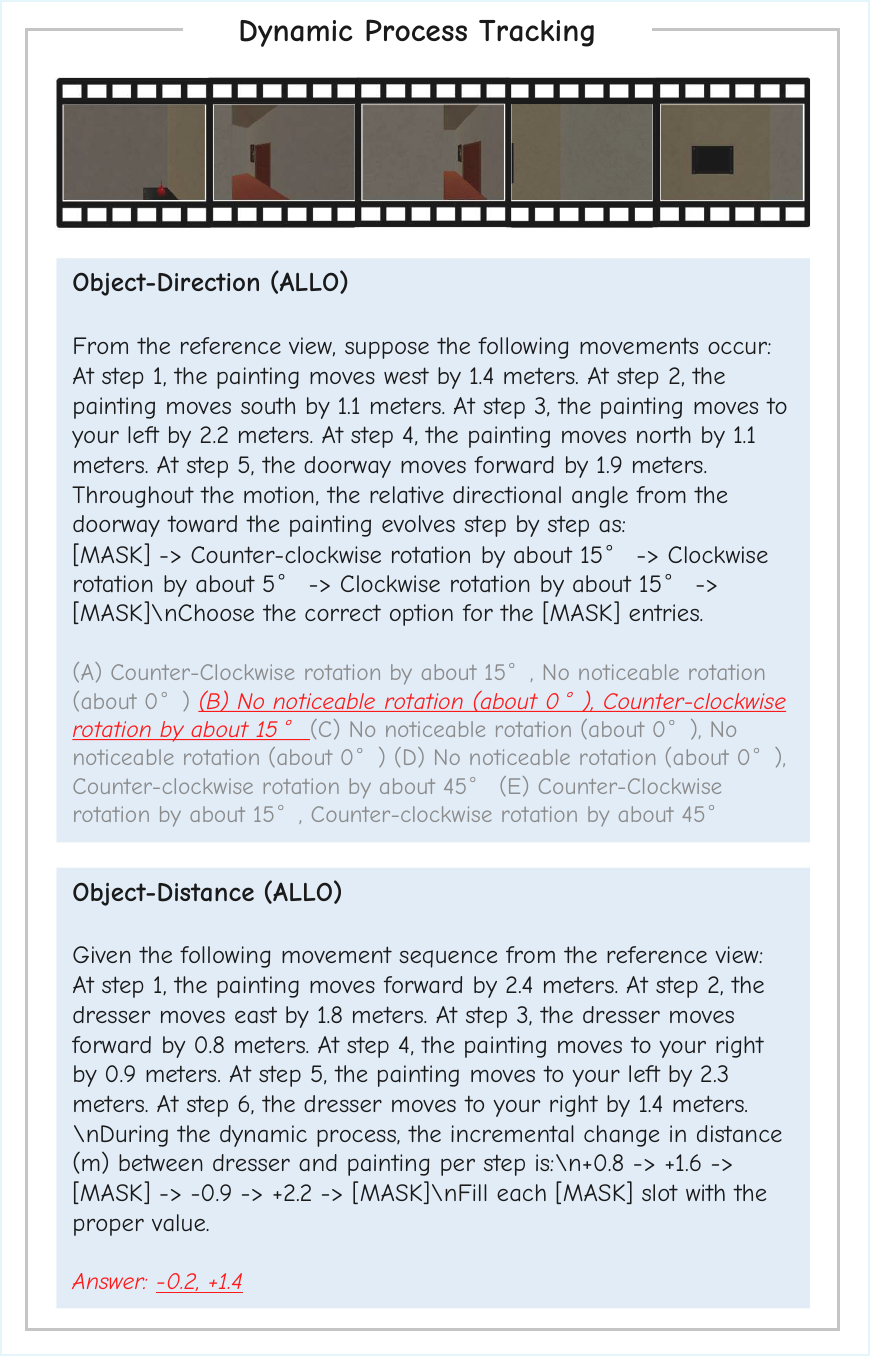}
    \caption{Examples in MentalGrid (Part 3).} \label{fig:case3}
\end{figure*}

Figures \ref{fig:case1}, \ref{fig:case2} and \ref{fig:case3} present further examples from MentalGrid, spanning a wide range of environments and task types.

\section{Details of MentalGrid Data Curation Pipeline} \label{append:data_curation}
The construction of the MentalGrid benchmark follows a rigorous four-stage pipeline designed to produce high-quality spatial simulation tasks.

\subsection{Details of Environment Synthesis}
The pipeline begins by loading realistic 3D environments and populating them with navigable agents and interactable objects. For each scene, the system identifies a valid traversable position by sampling random coordinates and verifying them through a pathfinding module. To ensure spatial diversity and avoid overlapping scenes, a minimum distance constraint of 2.0 m is enforced between consecutive sampling points. Once a location is secured, the agent captures a series of panoramic observations by rotating its camera across multiple viewpoints. The number of views and the degree of overlap between adjacent frames are randomized to simulate varied perceptual conditions. A critical step in this stage is scene purification, where any objects not visible from the initial set of viewpoints are programmatically removed from the environment. This ensures that the resulting scene metadata remains focused on the entities that the agent can perceive and reason about, maintaining a minimum object count for task complexity.

\subsection{Details of Query Construction}
Following environment synthesis, the pipeline establishes the objective of the spatial task. Two distinct entities are selected from the curated scene to serve as the subject and the reference point. To prevent geometric ambiguity and ensure the task is resolvable, the system enforces a minimum physical separation between these entities. Based on the task category, such as direction estimation or distance comparison, the pipeline selects a linguistic template from a library of natural language prompts. This stage also determines the reference frame by choosing between an egocentric perspective relative to the agent or an allocentric perspective relative to the global coordinate system.
The question templates for the generation of each task are listed in Table \ref{tab:template}.

\subsection{Details of Action Specialization}
Once the query is established, the pipeline simulates environmental changes to transform the static scene into a dynamic trajectory. A sequence of actions ranging from one to ten steps is generated, with a minimum of three steps enforced for tasks involving process tracking. To ensure the actions are relevant to the subsequent query, the simulation prioritizes the movement of the target entities involved in the task. This creates a direct causal link between the action sequence and the evolving spatial states. Both the agent movements and the object displacements are interleaved and then synthesized into a chronological narrative utilizing varied instructional prefixes to simulate diverse natural language commands.

\subsection{Details of Answer Generation}
The final stage computes the ground truth by tracking the state of the environment throughout the simulation. For orientation tasks, continuous angular values are mapped onto a set of discrete categories such as 15$^\circ$ or 90$^\circ$ turns. 
In process tracking scenarios, the pipeline records the state at every intermediate step. To evaluate simulation integrity, specific states within the sequence are masked, requiring the model to infer these hidden snapshots from the overall action flow. 
To facilitate robust evaluation, the pipeline generates hard negative options sampled from neighboring spatial categories or by reversing the direction of movement. 
This ensures that a correct answer reflects true spatial simulation rather than simple elimination or heuristic-based guessing.

\begin{table*}[ht]
\centering

\small
\renewcommand{\arraystretch}{1.6}
\begin{tabularx}{\textwidth}{l | l l X}
\textbf{Task} & \textbf{QueryType} & \textbf{Reference} & \textbf{Question Template} \\ \hline

\multirow{8}{*}{\textbf{SSP}} & \multirow{2}{*}{Object-Distance} & EGO & \textit{How far are you from the \textcolor{red}{\{target\_obj\}} in meters?} \\
 &  & ALLO & \textit{How far is the \textcolor{red}{\{target\_obj\}} from the \textcolor{red}{\{view\_obj\}} (in meters)?} \\ \cline{2-4}
 & \multirow{2}{*}{Distance-Object} & EGO & \textit{Which object lies farthest from your position? \textcolor{red}{\{options\}}} \\
 &  & ALLO & \textit{Which object lies farthest from the \textcolor{red}{\{target\_obj\}}? \textcolor{red}{\{options\}}} \\ \cline{2-4}
 & \multirow{2}{*}{Object-Direction} & EGO & \textit{In which direction is the \textcolor{red}{\{target\_obj\}} positioned relative to you? \textcolor{red}{\{options\}}} \\
 &  & ALLO & \textit{In which direction is the \textcolor{red}{\{target\_obj\}} located relative to the \textcolor{red}{\{view\_obj\}}? \textcolor{red}{\{options\}}} \\ \cline{2-4}
 & \multirow{2}{*}{Direction-Object} & EGO & \textit{Which object lies in your \textcolor{red}{\{direction\}}? \textcolor{red}{\{options\}}} \\
 &  & ALLO & \textit{Which object is located in the \textcolor{red}{\{direction\}} of the \textcolor{red}{\{target\_obj\}}? \textcolor{red}{\{options\}}} \\ \hline

\multirow{8}{*}{\textbf{FSP}} & \multirow{2}{*}{Object-Distance} & EGO & \textit{\textcolor{red}{\{action\_text\}} What is the final distance (meters) between you and the \textcolor{red}{\{target\_obj\}}?} \\
 &  & ALLO & \textit{\textcolor{red}{\{action\_text\}} What is the final distance (meters) between the \textcolor{red}{\{target\_obj\}} and the \textcolor{red}{\{view\_obj\}}?} \\ \cline{2-4}
 & \multirow{2}{*}{Distance-Object} & EGO & \textit{\textcolor{red}{\{action\_text\}} After the movements, which object ends up farthest from you? \textcolor{red}{\{options\}}} \\
 &  & ALLO & \textit{\textcolor{red}{\{action\_text\}} After the movements, which object ends up farthest from the \textcolor{red}{\{target\_obj\}}? \textcolor{red}{\{options\}}} \\ \cline{2-4}
 & \multirow{2}{*}{Object-Direction} & EGO & \textit{\textcolor{red}{\{action\_text\}} After the movements, in which direction does the \textcolor{red}{\{target\_obj\}} end up relative to you? \textcolor{red}{\{options\}}} \\
 &  & ALLO & \textit{\textcolor{red}{\{action\_text\}} After the movements, in which direction does the \textcolor{red}{\{target\_obj\}} end up relative to the \textcolor{red}{\{view\_obj\}}? \textcolor{red}{\{options\}}} \\ \cline{2-4}
 & \multirow{2}{*}{Direction-Object} & EGO & \textit{\textcolor{red}{\{action\_text\}} After the movements, which object ends up in your \textcolor{red}{\{direction\}}? \textcolor{red}{\{options\}}} \\
 &  & ALLO & \textit{\textcolor{red}{\{action\_text\}} After the movements, which object ends up in the \textcolor{red}{\{direction\}} of the \textcolor{red}{\{target\_obj\}}? \textcolor{red}{\{options\}}} \\ \hline

\multirow{8}{*}{\textbf{DPT}} & \multirow{2}{*}{Object-Distance} & EGO & \textit{\textcolor{red}{\{action\_text\}} During the movement sequence, the per-step change in distance (meters) between you and \textcolor{red}{\{target\_obj\}} is: \textcolor{red}{\{masked\_seq\}}} \\
 &  & ALLO & \textit{\textcolor{red}{\{action\_text\}} During the movement sequence, the per-step change in distance (meters) between \textcolor{red}{\{target\_obj\}} and \textcolor{red}{\{view\_obj\}} is: \textcolor{red}{\{masked\_seq\}}} \\ \cline{2-4}
 & \multirow{2}{*}{Distance-Object} & EGO & \textit{\textcolor{red}{\{action\_text\}} During the movement, the objects furthest from you, in sequence, are: \textcolor{red}{\{masked\_seq\}} \textcolor{red}{\{options\}}} \\
 &  & ALLO & \textit{\textcolor{red}{\{action\_text\}} During the movement, the objects furthest from the \textcolor{red}{\{target\_obj\}}, in sequence, are: \textcolor{red}{\{masked\_seq\}} \textcolor{red}{\{options\}}} \\ \cline{2-4}
 & \multirow{2}{*}{Object-Direction} & EGO & \textit{\textcolor{red}{\{action\_text\}} During the movement process, the changes in the relative direction angle from you to the \textcolor{red}{\{target\_obj\}} at each step are: \textcolor{red}{\{masked\_seq\}}} \\
 &  & ALLO & \textit{\textcolor{red}{\{action\_text\}} During the movement process, from your perspective, the changes in the relative direction angle from the \textcolor{red}{\{view\_obj\}} to the \textcolor{red}{\{target\_obj\}} at each step are: \textcolor{red}{\{masked\_seq\}}} \\ \cline{2-4}
 & \multirow{2}{*}{Direction-Object} & EGO & \textit{\textcolor{red}{\{action\_text\}} Over the course of the movement, the object that falls within your \textcolor{red}{\{direction\}}, in sequence, is: \textcolor{red}{\{masked\_seq\}}} \\
 &  & ALLO & \textit{\textcolor{red}{\{action\_text\}} During the movement process, the object occupying the \textcolor{red}{\{direction\}} of the \textcolor{red}{\{target\_obj\}}, in sequence, is: \textcolor{red}{\{masked\_seq\}}} \\ 
\end{tabularx}
\caption{Question Templates for tasks in MentalGrid.}\label{tab:template}
\end{table*}

\subsection{Details of Quality Control}
The reliability and rigor of the MentalGrid benchmark are sustained through a multi-layered quality control framework integrated into the data curation pipeline. During the environment synthesis phase, spatial diversity is guaranteed by enforcing a strict minimum distance constraint between sampling points, which ensures that each scene represents a unique sub-region and prevents data redundancy. 
Furthermore, the scene purification process programmatically eliminates occluded or distant objects that are absent from the agent's initial set of viewpoints, thereby ensuring that all entities involved in the subsequent simulation are properly grounded in the initial observation. 
To maintain the reasoning complexity of the tasks, samples that fail to meet a minimum visible object count are automatically discarded. 
In the action generation stage, the pipeline prioritizes the movement of queried objects through a targeted sampling ratio, which creates a robust causal link between the observed dynamics and the final spatial question. 
For task evaluation, the generation of hard negatives involves strategic sampling from neighboring spatial categories and direction reversals, which effectively prevents the model from relying on simple elimination or language-based heuristics. 
Finally, the use of diverse instructional prefixes and randomized reference frames ensures that the benchmark remains resilient to template-level over-fitting while maintaining high linguistic variability.

\section{Details of Main Evaluation Experiments } \label{append:main_eval}
\subsection{Experimental Setup}
We utilize the vLLM framework \citep{vllm} deployed on 8 NVIDIA RTX 3090 GPUs for the inference of all open-source models. To ensure reproducibility, we adopt a greedy decoding strategy for all models unless otherwise specified. Specifically, the temperature is set to 0 and both top-p and top-k are set to 1. The specific system prompts used are detailed in Figure \ref{fig:infer_prompt}.

\begin{figure*}[t]
\centering
\begin{tcolorbox}[
    title=System Prompt for Model Inference,
    colback=white,
    width=0.90\linewidth,
    colframe=frameblue!50,
    coltitle=gray!40!black,
    fonttitle=\bfseries,
    arc=1mm,
    boxrule=0.6mm,
    left=1mm,
    right=1mm,
    top=1mm,
    bottom=1mm,
]
\linespread{1.2}\selectfont
You are a spatial-reasoning assistant.\\
\\
\textbf{\texttt{<RULE>}}\\
1. You are an embodied observer represented by the camera.\\
2. You may receive EITHER: \\
\hspace*{1em} - a single image, OR  \\
\hspace*{1em} - multiple images captured at the SAME time step from different viewpoints of the SAME environment.\\
3. If only ONE image is provided: \\
\hspace*{1em} - Your position corresponds to the camera viewpoint of that image.\\
4. If MULTIPLE images are provided: \\ 
\hspace*{1em} - Image 1 defines the reference frame. \\
\hspace*{1em} - All other images provide complementary viewpoints and depict the same static moment.\\
5. At the initial reference frame, your facing direction equals the camera viewing direction of the reference image (Image 1) and is defined as North (+Y-axis) in a world-fixed coordinate system.\\
6. Relative directions (forward, backward, left, right) are always defined with respect to your CURRENT facing direction and rotate together with it.\\
\textbf{\texttt{</RULE>}}\\
\\
\textbf{\texttt{<TASK>}}\\
You will receive:\\
1. \textbf{Image(s)} - one or more RGB frames depicting the same scene at a single time step.\\
2. \textbf{Actions (Optional)} - a natural language sequence of actions describing the movements of you (i.e., the camera) and the objects. We ignore any blockages while moving.\\
3. \textbf{Question} - a natural-language query about spatial relationships between objects in the image.\\
4. \textbf{Options (Optional)} - 5 answer candidates, each tagged by a capital letter (A, B, C, D, E).\\
\textbf{\texttt{</TASK>}}\\
\\
\textbf{\texttt{<REQUIREMENTS>}}\\
- Always ground your answer in the visual evidence; do not hallucinate unseen objects.\\
- If uncertain, pick the most plausible option—never refuse or reply "insufficient information."\\
\textbf{\texttt{</REQUIREMENTS>}}\\
\end{tcolorbox}
\caption{The prompt used for model inference.}
\label{fig:infer_prompt}
\end{figure*}

\subsection{More Evaluation Results}
In this section, we provide a comprehensive set of evaluation results on our benchmark. This includes the full evaluation results across all question types as presented in Table \ref{tab:full_results}. Furthermore, we report the complete results on MentalGrid-tiny in Table \ref{tab:tiny} to facilitate a direct comparison with human performance.

\begin{table*}[h]
\centering

\resizebox{\linewidth}{!}{
\begin{tabular}{lccccccccccccc}
\toprule
\multirow{2.5}{*}{\textbf{Model}} & \multicolumn{4}{c}{\textbf{Static Spatial Perception}} & \multicolumn{4}{c}{\textbf{Final State Prediction}} & \multicolumn{4}{c}{\textbf{Dynamic Process Tracking}} & \multirow{2.5}{*}{\textbf{Overall}} \\ 
\cmidrule(lr){2-5} \cmidrule(lr){6-9} \cmidrule(lr){10-13}
 & \text{dir-obj} & \text{dis-obj} & \text{obj-dir} & \text{obj-dis} & \text{dir-obj} & \text{dis-obj} & \text{obj-dir} & \text{obj-dis} & \text{dir-obj} & \text{dis-obj} & \text{obj-dir} & \text{obj-dis} & \\ 
\midrule
GPT-5-mini & 42.57 & 54.20 & 42.98 & 38.54 & 41.53 & 43.51 & 55.87 & 43.55 & 27.02 & 42.24 & 66.08 & 41.39 & 45.70 \\
Gemini-3-Flash & 42.08 & 49.24 & 47.85 & 41.60 & 33.88 & 46.18 & 50.72 & 41.72 & 22.18 & 38.79 & 61.95 & 50.68 & 45.24 \\
GPT-4o & 43.56 & 47.71 & 32.66 & 35.62 & 25.68 & 35.50 & 38.11 & 27.42 & 19.35 & 29.31 & 16.52 & 22.24 & 30.85 \\
Gemma3-12B & 27.72 & 33.59 & 28.37 & 33.64 & 28.96 & 27.48 & 31.81 & 29.37 & 33.06 & 26.72 & 21.83 & 14.01 & 27.85 \\
InternVL3.5-8B & 29.70 & 39.69 & 36.39 & 30.97 & 27.87 & 22.90 & 34.38 & 26.45 & 20.56 & 25.86 & 21.24 & 16.93 & 27.84 \\
SpaceR-7B & 31.68 & 32.44 & 24.93 & 38.97 & 24.59 & 23.66 & 20.63 & 27.79 & 26.21 & 22.41 & 23.60 & 6.45 & 25.07 \\
Qwen2.5-VL-7B & 25.74 & 31.30 & 29.80 & 16.56 & 27.87 & 26.34 & 27.79 & 19.66 & 31.05 & 23.28 & 28.91 & 12.07 & 24.59 \\
Spatial-SSRL-7B & 28.71 & 35.50 & 26.36 & 26.22 & 17.49 & 24.43 & 26.65 & 17.62 & 21.77 & 22.41 & 22.42 & 7.57 & 22.87 \\
SpatialLadder-3B & 20.79 & 25.57 & 21.78 & 17.88 & 14.75 & 18.70 & 15.19 & 23.09 & 14.92 & 31.03 & 26.25 & 8.28 & 19.30 \\
SpatialThinker-3B & 28.22 & 26.34 & 26.07 & 15.10 & 25.68 & 23.28 & 26.36 & 3.04 & 18.15 & 23.28 & 12.68 & 3.46 & 18.10 \\
Qwen2.5-VL-3B & 18.32 & 28.63 & 18.62 & 10.72 & 13.11 & 18.32 & 22.35 & 2.35 & 6.85 & 7.76 & 3.54 & 2.93 & 12.54 \\ 
\bottomrule
\end{tabular}}
\caption{Evaluation results on each query type of MentalGrid.}
\label{tab:full_results}
\end{table*}

\begin{table*}[t]
\centering
\resizebox{\linewidth}{!}{
\begin{tabular}{lccccccc}
\toprule
\multirow{2}{*}{\textbf{Model}} & \textbf{SSP} & \multicolumn{2}{c}{\textbf{FSP}} & \multicolumn{2}{c}{\textbf{DPT}} & \multicolumn{2}{c}{\textbf{Overall}} \\ 
\cmidrule(r){2-2} \cmidrule(r){3-4} \cmidrule(r){5-6} \cmidrule(r){7-8} 
&  Acc (\%) \textcolor{textgreen}{$\boldsymbol{\uparrow}$}   & Acc (\%) \textcolor{textgreen}{$\boldsymbol{\uparrow}$}  & Drop (\%) \textcolor{textred}{$\boldsymbol{\downarrow}$}  & Acc (\%) \textcolor{textgreen}{$\boldsymbol{\uparrow}$}   & Drop (\%) \textcolor{textred}{$\boldsymbol{\downarrow}$} & Acc (\%) \textcolor{textgreen}{$\boldsymbol{\uparrow}$}  & Drop (\%) \textcolor{textred}{$\boldsymbol{\downarrow}$} \\
\midrule

Human Level & 71.80 & 76.95 & -7.17 & 76.00 & -5.85 & 74.73 & -6.51\\
Gemini-3.1-Pro   & 64.00 & 65.40 &  -2.19 & 47.40 &  25.94 & 58.93 &  11.88 \\
Gemini-3-Flash & 60.00 & 52.00 & 13.33 & 51.50 & 14.17 & 54.50 & 13.75 \\
GPT-5  & 53.60 & 38.40 &  28.36 & 59.70 & -11.38 & 50.57 &   8.49 \\
GPT-4o & 50.00 & 35.00 & 30.00 & 18.70 & 62.60 & 34.57 & 46.30 \\
Gemma3-12B & 38.40 & 34.40 & 10.42 & 21.60 & 43.75 & 31.47 & 27.08 \\
InternVL3.5-8B & 50.80 & 27.80 & 45.28 & 15.40 & 69.69 & 31.33 & 57.48 \\
SpaceR-7B & 35.40 & 21.20 & 40.11 & 20.10 & 43.22 & 25.57 & 41.67 \\
Qwen2.5-VL-7B & 30.20 & 15.00 & 50.33 & 26.80 & 11.26 & 24.00 & 30.79 \\
Spatial-SSRL-7B & 29.20 & 14.40 & 50.68 & 20.00 & 31.51 & 21.20 & 41.10 \\
SpatialLadder-3B & 21.60 & 16.60 & 23.15 & 13.80 & 36.11 & 17.33 & 29.63 \\
SpatialThinker-3B & 17.00 & 14.40 & 15.29 & 14.80 & 12.94 & 15.40 & 14.12 \\
Qwen2.5-VL-3B & 15.80 & 16.40 & -3.80 & 4.00 & 74.68 & 12.07 & 35.44 \\
\bottomrule
\end{tabular}}
\caption{More evaluation results on MentalGrid-tiny.} 
\label{tab:tiny}

\end{table*}

\section{Details of Perception Complexity Experiments}\label{append:perception_exp}
\begin{figure*}[h]
\centering
\includegraphics[width=\textwidth]{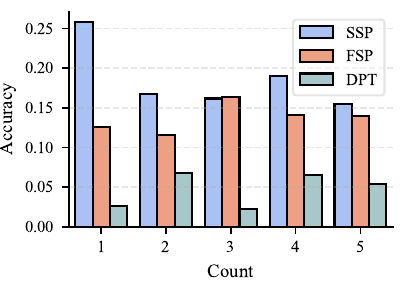}
\caption{Evaluation under different image count on Qwen2.5-VL-3B.} \label{fig:img_cnt}
\end{figure*}

\begin{figure*}[h]
  \centering
  \begin{subfigure}[t]{0.333\textwidth}
    \centering
    \includegraphics[width=\linewidth]{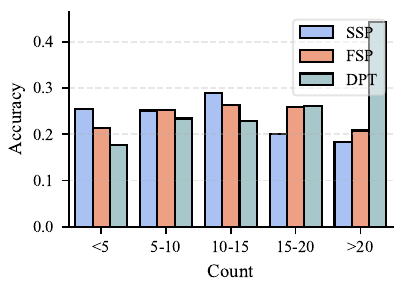}
    \caption{Image Count}
  \end{subfigure}\hfill
  \begin{subfigure}[t]{0.333\textwidth}
    \centering
    \includegraphics[width=\linewidth]{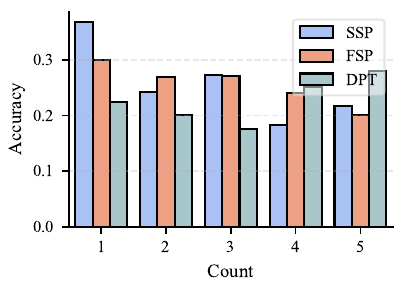}
    \caption{Object Count}
  \end{subfigure}\hfill
  \begin{subfigure}[t]{0.333\textwidth}
    \centering
    \includegraphics[width=\linewidth]{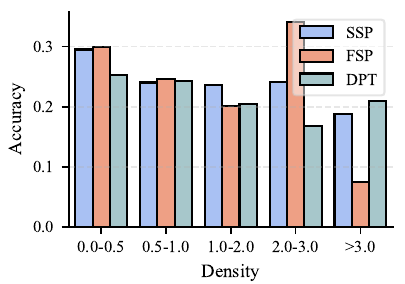}
    \caption{Spatial Density}
  \end{subfigure}
  \caption{Comparison of methods under different perception complexity on Qwen2.5-VL-7B.}
  \label{fig:perception_7b}
\end{figure*}

\begin{figure*}[h]
  \centering
  \begin{subfigure}[t]{0.333\textwidth}
    \centering
    \includegraphics[width=\linewidth]{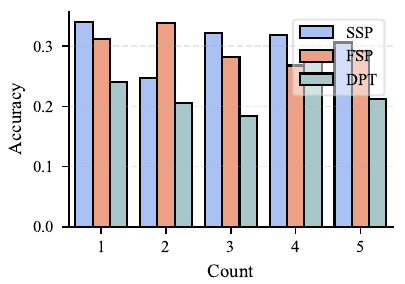}
    \caption{Image Count}
  \end{subfigure}\hfill
  \begin{subfigure}[t]{0.333\textwidth}
    \centering
    \includegraphics[width=\linewidth]{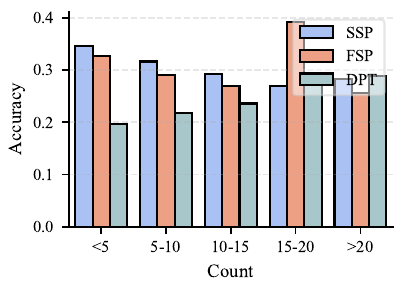}
    \caption{Object Count}
  \end{subfigure}\hfill
  \begin{subfigure}[t]{0.333\textwidth}
    \centering
    \includegraphics[width=\linewidth]{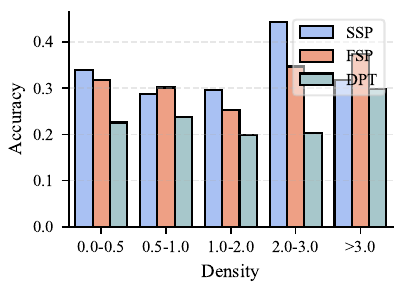}
    \caption{Spatial Density}
  \end{subfigure}
  \caption{Comparison of methods under different perception complexity on Gemma3-12B.}
  \label{fig:perception_gemma}
\end{figure*}

\begin{figure*}[h]
  \centering
  \begin{subfigure}[t]{0.333\textwidth}
    \centering
    \includegraphics[width=\linewidth]{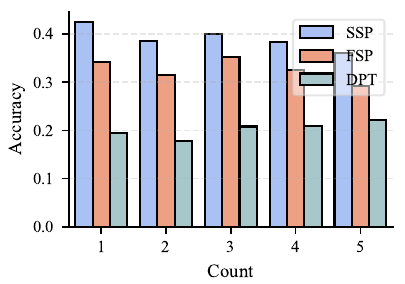}
    \caption{Image Count}
  \end{subfigure}\hfill
  \begin{subfigure}[t]{0.333\textwidth}
    \centering
    \includegraphics[width=\linewidth]{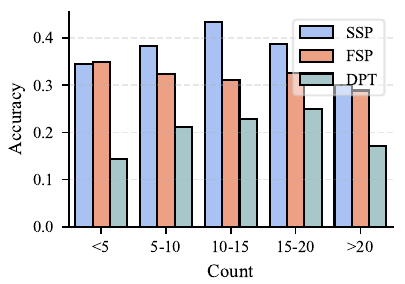}
    \caption{Object Count}
  \end{subfigure}\hfill
  \begin{subfigure}[t]{0.333\textwidth}
    \centering
    \includegraphics[width=\linewidth]{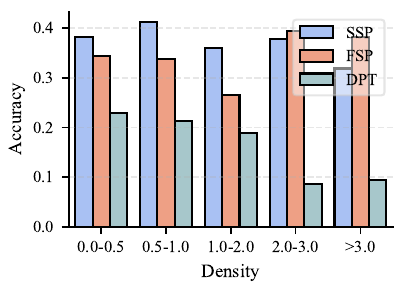}
    \caption{Spatial Density}
  \end{subfigure}
  \caption{Comparison of methods under different perception complexity on GPT-4o.}
  \label{fig:perception_4o}
\end{figure*}

Here, we introduce the image count as a new factor to reflect perceptual complexity.
As illustrated in Figure \ref{fig:img_cnt}, when fewer input images are provided (e.g., a single image), models achieve noticeably higher performance on static spatial perception. 
Conversely, performance on spatial simulation tasks (FSP and DPT) does not exhibit a corresponding improvement as the number of input images decreases. This suggests that the difficulty of mental simulation is decoupled from the density of the visual input stream.
Moreover, we extended the analytical experiments from \S \ref{sec:4.3} to  Qwen2.5-VL-7B, Gemma3-12B, and GPT-4o. The results, reported in Figure \ref{fig:perception_7b}, \ref{fig:perception_gemma}, and \ref{fig:perception_4o}, provide further empirical support for the conclusions presented in the main text.

\begin{figure*}[t]
  \centering
  \begin{subfigure}[t]{0.49\linewidth}
    \centering
    \includegraphics[width=\linewidth]{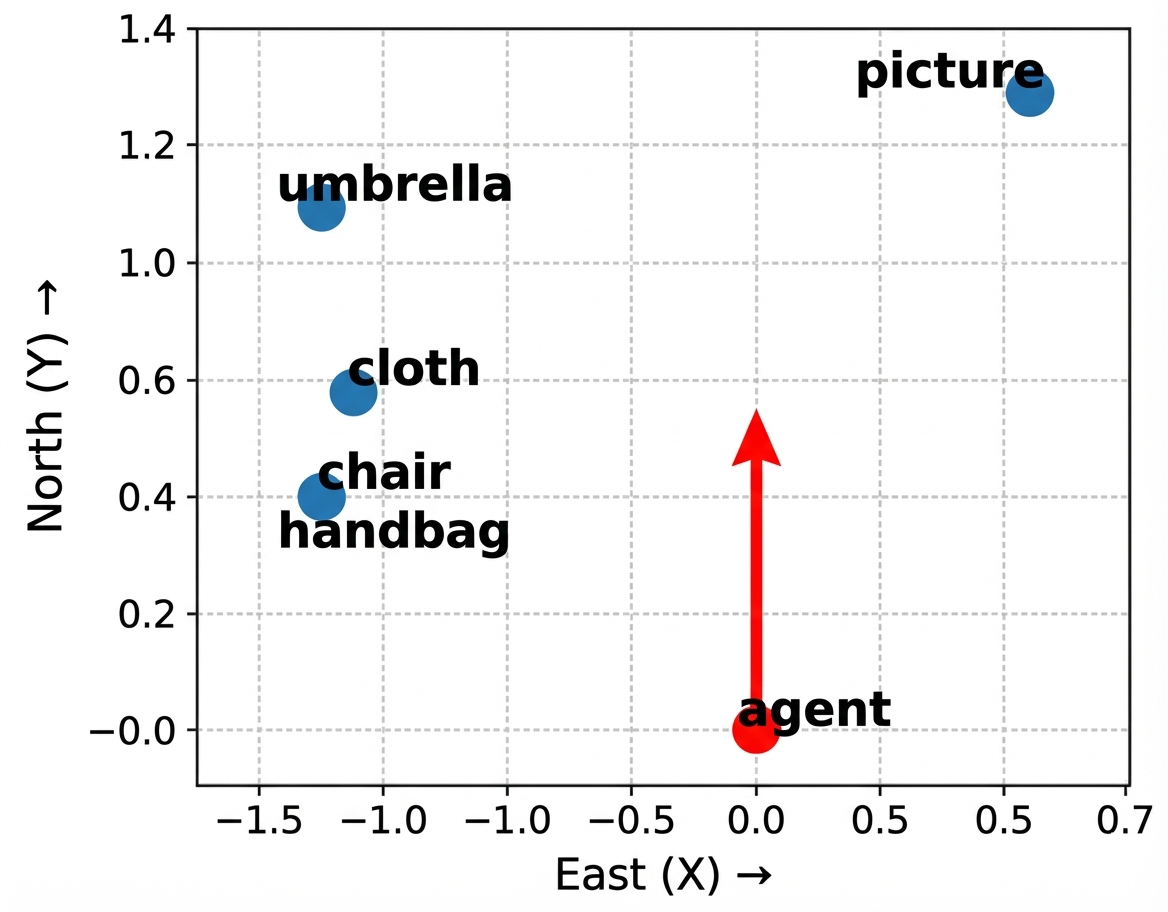}
    \caption{2D Map Image} \label{fig:map_image_format}
  \end{subfigure}\hfill
  \begin{subfigure}[t]{0.49\linewidth}
    \centering
    \includegraphics[width=\linewidth]{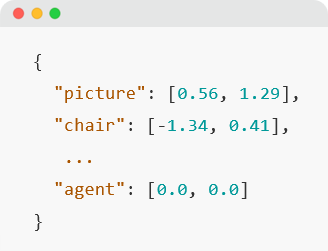}
    \caption{2D Map Text} \label{fig:map_text_format}
  \end{subfigure}
  \caption{Two simplified input formats for perception.}
  \label{fig:input_format}
\end{figure*}

\section{Details of Simulation Complexity Experiments}\label{append:simulation_exp}
\begin{figure*}[h]
  \centering
  \begin{subfigure}[t]{0.333\textwidth}
    \centering
    \includegraphics[width=\linewidth]{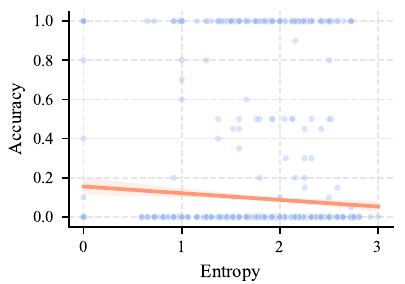}
    \caption{Raw Image} \label{fig:3d_scene_ent}
  \end{subfigure}\hfill
  \begin{subfigure}[t]{0.333\textwidth}
    \centering
    \includegraphics[width=\linewidth]{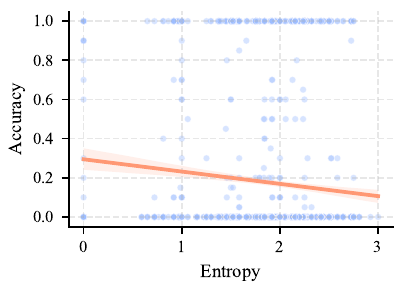}
    \caption{2D Map Image} \label{fig:image_map_ent}
  \end{subfigure}\hfill
  \begin{subfigure}[t]{0.333\textwidth}
    \centering
    \includegraphics[width=\linewidth]{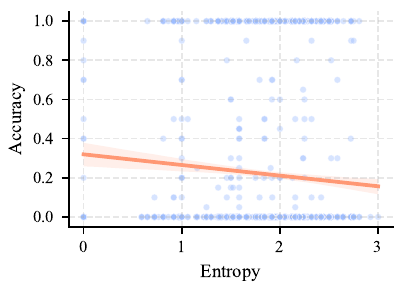}
    \caption{2D Map Text} \label{fig:text_map_ent}
  \end{subfigure}
  \caption{Evaluation under different action entropy on Qwen2.5-VL-3B.}
  \label{fig:sim_complex_ent}
\end{figure*}

\begin{figure*}[h]
  \centering
  \begin{subfigure}[t]{0.333\textwidth}
    \centering
    \includegraphics[width=\linewidth]{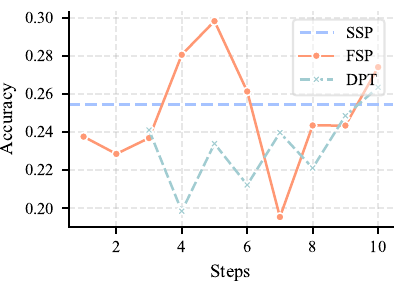}
    \caption{Raw Image} \label{fig:3d_scene}
  \end{subfigure}\hfill
  \begin{subfigure}[t]{0.333\textwidth}
    \centering
    \includegraphics[width=\linewidth]{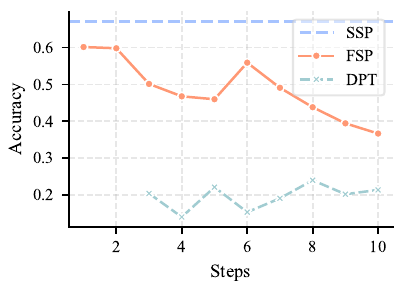}
    \caption{2D Map Image} \label{fig:image_map}
  \end{subfigure}\hfill
  \begin{subfigure}[t]{0.333\textwidth}
    \centering
    \includegraphics[width=\linewidth]{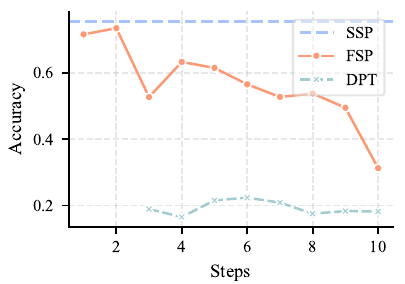}
    \caption{2D Map Text} \label{fig:text_map}
  \end{subfigure}
  \caption{Evaluation under different action steps on Qwen2.5-VL-7B.}
  \label{fig:sim_complex_7b}
\end{figure*}

\begin{figure*}[h]
  \centering
  \begin{subfigure}[t]{0.333\textwidth}
    \centering
    \includegraphics[width=\linewidth]{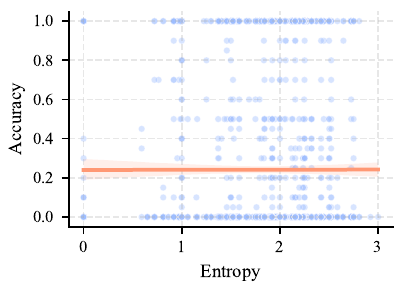}
    \caption{Raw Image} \label{fig:3d_scene_ent}
  \end{subfigure}\hfill
  \begin{subfigure}[t]{0.333\textwidth}
    \centering
    \includegraphics[width=\linewidth]{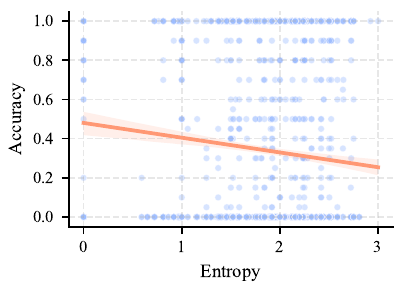}
    \caption{2D Map Image} \label{fig:image_map_ent}
  \end{subfigure}\hfill
  \begin{subfigure}[t]{0.333\textwidth}
    \centering
    \includegraphics[width=\linewidth]{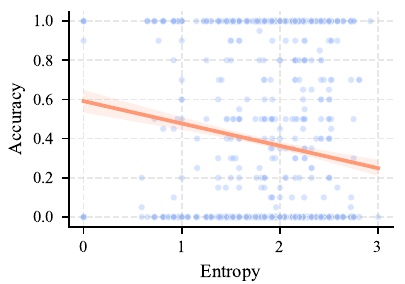}
    \caption{2D Map Text} \label{fig:text_map_ent}
  \end{subfigure}
  \caption{Evaluation under different action entropy on Qwen2.5-VL-7B.}
  \label{fig:sim_complex_ent_7b}
\end{figure*}

In addition to the number of action steps, we introduce Action Entropy as a metric to quantify the complexity and diversity of the simulation. For a given sequence of actions $\mathcal{A}$, we define each unique operator based on the combination $o$ of its reference frame and movement direction. Let $\mathcal{O}$ be the set of unique operators occurring in a trajectory of length $N$. The probability of an operator $o \in \mathcal{O}$ is calculated as $p(o) = \frac{count(o)}{L}$. The action entropy is then defined as:$$H(A) = -\sum_{o \in \mathcal{O}} p(o) \log_2 p(o)$$
A higher entropy indicates a more varied and less predictable sequence of spatial transitions, representing a more challenging simulation scenario. We report the accuracy of Qwen2.5-VL-3B under different input configurations as a function of action entropy. As illustrated in Figure \ref{fig:sim_complex_ent}, when coordinate-based inputs are introduced, models achieve strong spatial simulation performance at low action entropy, but exhibit a pronounced degradation as entropy increases. This trend is consistent with the results reported in \S \ref{sec:4.4}, further supporting our findings. We repeat this experiment on Qwen2.5-VL-7B to further demonstrate the generalizability of our findings (see Figure \ref{fig:sim_complex_7b}, \ref{fig:sim_complex_ent_7b}).

\section{Details of Spatial Simulation Training} \label{append:training}
\subsection{Details of Training Data Construction}
The specific system prompts used for CoT generation are detailed in Figure \ref{fig:cot_prompt}.

\begin{figure*}[t]
\centering
\scalebox{0.85}{
\begin{tcolorbox}[
    title=System Prompt for CoT Generation,
    colback=white,
    width=0.90\linewidth,
    colframe=frameblue!50,
    coltitle=gray!40!black,
    fonttitle=\bfseries,
    arc=1mm,
    boxrule=0.6mm,
    left=1mm,
    right=1mm,
    top=1mm,
    bottom=1mm,
]
\linespread{1.2}\selectfont
You are a spatial-reasoning assistant.\\
\\
\textbf{\texttt{<RULE>}}\\
1. You are an embodied observer represented by the camera.\\
2. You may receive EITHER: \\
\hspace*{1em} - a single image, OR  \\
\hspace*{1em} - multiple images captured at the SAME time step from different viewpoints of the SAME environment.\\
3. If only ONE image is provided: \\
\hspace*{1em} - Your position corresponds to the camera viewpoint of that image.\\
4. If MULTIPLE images are provided: \\ 
\hspace*{1em} - Image 1 defines the reference frame. \\
\hspace*{1em} - All other images provide complementary viewpoints and depict the same static moment.\\
5. At the initial reference frame, your facing direction equals the camera viewing direction of the reference image (Image 1) and is defined as North (+Y-axis) in a world-fixed coordinate system.\\
6. Relative directions (forward, backward, left, right) are always defined with respect to your CURRENT facing direction and rotate together with it.\\
\textbf{\texttt{</RULE>}}\\
\\
\textbf{\texttt{<TASK>}}\\
You will receive:\\
1. \textbf{Image(s)} - one or more RGB frames depicting the same scene at a single time step.\\
2. \textbf{Actions (Optional)} - a natural language sequence of actions describing the movements of you (i.e., the camera) and the objects. We ignore any blockages while moving.\\
3. \textbf{Question} - a natural-language query about spatial relationships between objects in the image.\\
4. \textbf{Options (Optional)} - 5 answer candidates, each tagged by a capital letter (A, B, C, D, E).\\
5. \textbf{Hint} - Initial object coordinates (Origin: camera in reference image; +Y: viewing direction; +X: camera's right).\\
\textbf{\texttt{</TASK>}}\\

\textbf{\texttt{<REQUIREMENTS>}}\\
- You SHOULD fully utilize the information in hint to form accurate internal spatial representations and improve reasoning quality. \\
- During reasoning and explanation, you MUST behave as if no such hints were provided. Do NOT mention, reference, or allude to hints. \\
- All reasoning must be expressed as if it were derived naturally from visual perception and embodied spatial understanding. \\
- Always ground your answer in the visual evidence; do not hallucinate unseen objects.\\
\textbf{\texttt{</REQUIREMENTS>}}\\
\end{tcolorbox}}
\caption{The prompt used for CoT generation.}
\label{fig:cot_prompt}
\end{figure*}

\subsection{Details of Training Implementation}
For Stage 1 and Stage 2, we implement SFT using the LLaMA-Factory framework \citep{llamafactory}. The models are optimized with the hyperparameter configurations in Table \ref{tab:sft_config}.
For Stage 3, we implement GRPO using the Verl framework \citep{verl}. The corresponding reinforcement learning configurations are provided in Table \ref{tab:rl_config}. All experiments are conducted on 4 NVIDIA A100 GPUs.

\begin{table}[h]
    \centering
    
    \begin{tabular}{l l}
        \toprule
        \textbf{Hyperparameter} & \textbf{Setting} \\
        \midrule
        Optimizer & AdamW \\
        Learning rate & $1 \times 10^{-5}$ \\
        LR scheduler & Cosine \\
        Max sequence length & 8192 \\
        Batch size per device & 4 \\
        Gradient accumulation steps & 4 \\
        Training epochs & 2.0 \\
        Warmup ratio & 0.1 \\
        Mixed precision & BF16 \\
        DeepSpeed strategy & ZeRO-3 \\
        \midrule
        \textbf{Trainable Modules} & \textbf{Status} \\
        Vision Tower & Frozen \\
        Multi-modal Projector & Frozen \\
        Language Model & Full Fine-tuning \\
        \bottomrule
    \end{tabular}
    \caption{Hyperparameters for Stage 1 and Stage 2 (SFT).}
    \label{tab:sft_config}
\end{table}

\begin{table}[h]
    \centering
   
    \begin{tabular}{l l}
        \toprule
        \textbf{Hyperparameter} & \textbf{Setting} \\
        \midrule
        Algorithm & GRPO \\
        Learning rate  & $1 \times 10^{-6}$ \\
        Group size  & 4 \\
        KL divergence coefficient & 0.01 \\
        KL loss type & Low-variable KL \\
        Entropy coefficient & 0 \\
        Max prompt length & 8192 \\
        Max response length & 1024 \\
        Training batch size & 64 \\
        PPO mini-batch size & 16 \\
        Micro-batch size per GPU & 1 \\
        Total epochs & 2.0 \\
        Mixed precision & BF16 \\
        Parallelism strategy & FSDP + TP ($size=2$) \\
        \bottomrule
    \end{tabular}
     \caption{Hyperparameters for Stage 3 (RL).}
    \label{tab:rl_config}
\end{table}

\section{Additional Spatial Simulation Comparison}\label{append:method_eval}
We reproduce the MindJourney-SVC and MindJourney-SWM methods as inference-time world-model augmentation approaches on MentalGrid-tiny, using GPT-4o as the backbone model \citep{MindJourney}. The results in Table \ref{tab:mindjourney} indicate that, although methods like MindJourney incur significantly lower computational cost compared to our training approach, their performance falls far short of what is achieved with our curriculum training. Based on comparisons with direct CoT reasoning using GPT-4o, we conclude that in our more complex spatial reasoning tasks, the quality of images generated by existing world models is insufficient, introducing substantial noise that hinders model reasoning.

\begin{table}[t]                                            
  \centering
  \resizebox{\linewidth}{!}{                                                                                                                                     
  \begin{tabular}{lcccc}
  \toprule                                                                                                                                                       
  \textbf{Model} & \textbf{SSP} & \textbf{FPT} & \textbf{DPT} & \textbf{All} \\
  \midrule                                                                                                                                                       
  MindJourney-SVC & 36.60 & 30.80 & 22.80 & 30.10 \\
  MindJourney-SWM & 37.20 & 29.40 & 16.00 & 27.50 \\                                                                                                             
  GPT-4o          & \underline{50.00} & \underline{35.00} & \underline{18.70} & \underline{34.60} \\                                                             
  \midrule                                                                                                                                                       
  \textbf{SpatialWalker} & \textbf{52.00} & \textbf{53.20} & \textbf{41.70} & \textbf{49.00} \\                                                                                                                                           
  \bottomrule                                                                                                                                                    
  \end{tabular}                                                                                                                                                  
  }               
  \caption{Evaluation results on MentalGrid-tiny. Best results are \textbf{bolded} and second best are \underline{underlined}.}
  \label{tab:mindjourney}                                                                                                                                        
  \end{table}

\begin{table*}[t]                                                                                                                                              
  \centering                                                                                                                                                     
  \resizebox{\linewidth}{!}{                                                                                                                                     
  \begin{tabular}{lcccccccccccc}                                                                                                                                 
  \toprule             
  \multirow{2}{*}{\textbf{Model}} & \multicolumn{3}{c}{\textbf{MulSeT}} & \multicolumn{8}{c}{\textbf{SpinBench}} \\
  \cmidrule(r){2-4} \cmidrule(r){5-12}                                                                                                                           
  & OR. & DC. & AT. & CVS. & IM. & DR. & PT. & ORG. & DT. & MR. & All \\
  \midrule                                                                                                                                                       
 Qwen2.5-VL-3B & 10.02 & 27.00 & 3.29 & 34.36 & 27.41 & \textbf{42.78} & 43.23 & 49.06 & 50.00 & 19.23 & 40.59 \\                           

\textbf{SpatialWalker} & \textbf{26.47} & \textbf{42.15} & \textbf{33.72} & \textbf{46.65} & \textbf{46.91} & 26.91 & \textbf{46.98} &     
  \textbf{82.55} & \textbf{52.56} & 19.23 & \textbf{52.40} \\                                                                                                    
\textbf{\textit{\textcolor{textgreen}{\quad $\Delta$ Improvement}}} & \textcolor{textgreen}{+16.45} & \textcolor{textgreen}{+15.15} &      
  \textcolor{textgreen}{+30.43} & \textcolor{textgreen}{+12.29} & \textcolor{textgreen}{+19.50} & \textcolor{textred}{-15.87} & \textcolor{textgreen}{+3.75} &       
  \textcolor{textgreen}{+33.49} & \textcolor{textgreen}{+2.56} & \textcolor{textgreen}{+0.00} & \textcolor{textgreen}{+11.81} \\
  \bottomrule                                                                                                                                                    
  \end{tabular}   
  }
  \caption{Evaluation results on MulSeT and SpinBench. Best results are \textbf{bolded}.}
  \label{tab:mulset_spinbench}                                                                                                                                   
  \end{table*}     

\section{Additional Out-of-domain Benchmark Evaluation}\label{append:ood}
We further compare SpatialWalker with its base model, Qwen2.5-VL-3B, on two other spatial reasoning benchmarks: MulSeT\citep{MulSeT} and SpinBench \citep{SpinBench}. The results are demonstrated in Table \ref{tab:mulset_spinbench}, which show that our method maintains strong generalization across a broader range of benchmarks.

\section{Additional Ablation Study}\label{append:ablation}
\begin{figure*}[h]
    \centering
    \includegraphics[width=0.9\linewidth]{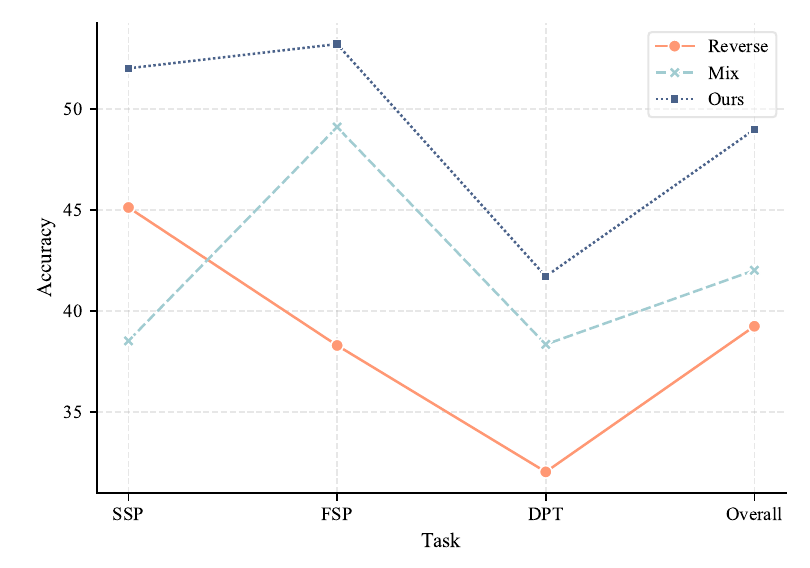}
    \caption{Additional ablation study results.} \label{fig:ablation_append}
\end{figure*}

To further validate the effectiveness of our proposed multi-stage training strategy, we conduct two additional ablation studies: \textbf{(1) Reverse:} We swapped the execution order of Stage 1 and Stage 2 to investigate the necessity of our sequential training curriculum.
\textbf{(2) Mix:} We integrated the training objectives of both Stage 1 and Stage 2 into a single-stage joint training process to compare against our decoupled design.
The experimental results in Figure \ref{fig:ablation_append} indicate that performance declines across all tasks under both ablation settings, demonstrating the necessity of our progressive design.

\end{document}